\documentclass[11pt]{amsart}
\def\isdraft{0}

\usepackage[dvipsnames]{xcolor}
\usepackage{bbm}
\usepackage{graphicx}
\usepackage{amsaddr} 
\usepackage{mathtools}
\usepackage{amsmath,amssymb,amsfonts,dsfont}
\usepackage{prettyref} 
\usepackage[utf8]{inputenc}
\usepackage{enumitem}
\usepackage[hyphens]{url} 
\usepackage{tikz-cd}
\usepackage{tikz}
\usetikzlibrary{positioning}
\usetikzlibrary{arrows}
\usepackage{bussproofs}
\usepackage[mathscr]{euscript}
\usepackage{hyperref}
\usepackage{amsthm}
\usepackage{a4wide}
\usepackage[color=black,textcolor=white\if\isdraft0,disable\fi]{todonotes}
\usepackage{stackengine}
\usepackage{stmaryrd}
\usepackage{wrapfig}
\usepackage{pdfpages}
\usepackage{marginnote}
\usepackage{float}
\graphicspath{{fig/}}
\usetikzlibrary{arrows,automata,topaths,matrix,positioning,fit}
\usepgflibrary{shapes.geometric}
\usetikzlibrary{shapes.geometric}
\tikzset{every state/.style={minimum size=0pt}}
\usepackage{footnote}
\usepackage{float}
\usepackage{tabularx}
\usepackage{txfonts}
\usepackage[normalem]{ulem}
\usepackage{pifont}
\usepackage[most]{tcolorbox}
\usepackage{shuffle}

\newtheorem{theorem}{Theorem}

\newtheorem{fact}[theorem]{Fact}
\newtheorem{lemma}[theorem]{Lemma}
\newtheorem{proposition}[theorem]{Proposition}

\theoremstyle{definition} 

\newtheorem{definition}[theorem]{Definition}

\newtheorem{example}[theorem]{Example}

\newrefformat{§}{§\ref{#1}}
\newrefformat{a}{Answer \ref{#1}}
\newrefformat{ana}{Analogy \ref{#1}}
\newrefformat{algorithm}{Algorithm \ref{#1}}
\newrefformat{appendix}{Appendix \ref{#1}}
\newrefformat{claim}{Claim \ref{#1}}
\newrefformat{conclusion}{Conclusion \ref{#1}}
\newrefformat{convention}{Convention \ref{#1}}
\newrefformat{cj}{Conjecture \ref{#1}}
\newrefformat{c}{Corollary \ref{#1}}
\newrefformat{q}{(\ref{#1})}
\newrefformat{e}{Example \ref{#1}}
\newrefformat{exe}{Exercise \ref{#1}}
\newrefformat{d}{Definition \ref{#1}}
\newrefformat{Done}{Done \ref{#1}}
\newrefformat{f}{Fact \ref{#1}}
\newrefformat{fig}{Figure \ref{#1}}
\newrefformat{h}{Hypothesis \ref{#1}}
\newrefformat{idea}{Idea \ref{#1}}
\newrefformat{i}{Item \ref{#1}}
\newrefformat{l}{Lemma \ref{#1}}
\newrefformat{note}{Note \ref{#1}}
\newrefformat{n}{Notation \ref{#1}}
\newrefformat{o}{Observation \ref{#1}}
\newrefformat{problem}{Problem \ref{#1}}
\newrefformat{p}{Proposition \ref{#1}}
\newrefformat{pseudocode}{Pseudocode \ref{#1}}
\newrefformat{Q}{Q. \ref{#1}}
\newrefformat{r}{Remark \ref{#1}}
\newrefformat{table}{Table \ref{#1}}
\newrefformat{t}{Theorem \ref{#1}}
\newrefformat{w}{Warning \ref{#1}}

\newcommand{\boxblacktriangle}{\mathrel{\ooalign{$\square$\cr\kern0.07ex\hbox{\scalebox{0.9}{$\blacktriangle$}}}}}
\newcommand{\boxtriangle}{\mathrel{\ooalign{$\square$\cr\kern0.07ex\hbox{\scalebox{0.9}{$\triangle$}}}}}

\title{
	Generalization-based similarity
}
\author{
	Christian Anti\'c
}
\address{
	christian.antic@icloud.com\\
	Vienna University of Technology\\
	Vienna, Austria
}

\begin{document}

\begin{abstract} 
	Detecting and exploiting similarities between seemingly distant objects is without doubt an important human ability. This paper develops \textit{from the ground up} an abstract algebraic and qualitative notion of similarity based on the observation that sets of generalizations encode important properties of elements. We show that similarity defined in this way has appealing mathematical properties. As we construct our notion of similarity from first principles using only elementary concepts of universal algebra, to convince the reader of its plausibility, we show that it can model fundamental relations occurring in mathematics and be naturally embedded into first-order logic via model-theoretic types. Finally, we sketch some potential applications to theoretical computer science and artificial intelligence.
\end{abstract}

\maketitle


\section{Introduction}\label{§:I}

Similarity is vital to many fields like, for example, data analysis and clustering \cite{Keeney76}, granular computing \cite{Yao99,Zadeh97}, approximate reasoning \cite{Dubois94,Esteva94}, and approximate information retrieval \cite{Klir95}.

Analogy-making is at the core of human and artificial intelligence \cite{Gust08,Hofstadter01,Hofstadter13,Krieger03,Polya54}, and detecting and exploiting similarities between seemingly distant objects is at the core of analogy-making, especially in analogical transfer \cite{Badra18} and case-based prediction \cite{Badra23}.

This paper therefore aims to give a mathematically precise answer to the (philosophical) question
\begin{quote} 
	\textit{\textbf{What is (the essence of) similarity?}}
\end{quote} 

There are essentially two ways of describing and interpreting the notion of similarity, namely in terms of a quantitative similarity measure or in terms of a binary relation which is reflexive and symmetric called a compatibility or tolerance relation \cite{Nehring97,Yao00}.

In this paper, we choose the latter path. That is, the purpose of this paper is to introduce \textit{from first principles} an abstract algebraic, \textit{qualitative}, and \textit{justification-based} notion of similarity based on sets of generalizations as motivated by the following observations. 

We say that a term $s(x_1,\ldots,x_n)$ is a \textit{\textbf{generalization}} of a natural number $a\in\mathbb N$ iff there is some sequence of natural numbers $o_1,\ldots,o_n$ such that $a=s(o_1,\ldots,o_n)$. For example, the term $2x$ is a generalization of $6$ since $6=2\cdot 3$. More generally speaking, the term $2x$ is a generalization of every even number, that is,
\begin{align*} 
	\text{$2x$ is a generalization of $a$} \quad\Leftrightarrow\quad a\text{ is even}.
\end{align*} We also have
\begin{align*} 
	\text{$x^2$ is a generalization of $a$} \quad\Leftrightarrow\quad a\text{ is a square number}.
\end{align*} 

These examples indicate that generalizations can encode important properties of elements, which motivates the following definition of generalization-based similarity. 

Let $\mathfrak A$ and $\mathfrak B$ be algebras over some joint language of algebras $L$, and let $a$ and $b$ be elements of the universes of $\mathfrak A$ and $\mathfrak B$, respectively. 

We denote the set of all generalizations of $a$ and $b$ by $\uparrow_ \mathfrak A a$ and $\uparrow_\mathfrak B b$, respectively. Since these sets of generalizations encode properties of $a$ and $b$ as noted above, the intersection
\begin{align*} 
	a\uparrow_ \mathfrak{(A,B)} b := (\uparrow_ \mathfrak A a)\cap (\uparrow_\mathfrak B b)
\end{align*} contains the \textit{joint properties} of $a$ and $b$ (since $\mathfrak A$ and $\mathfrak B$ have the same underlying language, the terms in both sets contain symbols from the same alphabet). 

We now say that $a$ and $b$ are \textit{\textbf{similar}} if this set of joint properties $a\uparrow_ \mathfrak{(A,B)} b$ is maximal with respect to $a$ and $b$ (see \prettyref{d:approx} for details). That is, we formalize a notion of maximal similarity \cite{Falklenhainer89,Gentner83}. What is essential here is that $a$ and $b$ can be from different algebras. 

It turns out that similarity defined in this way has appealing mathematical properties which we shall now discuss. 

The First \ref{t:FIT} and Second Isomorphism Theorems \ref{t:SIT} show that similarity is compatible with bijective structure-preserving mappings as desired. \prettyref{e:H_1}, on the other hand, shows that similarity and homomorphisms may not be compatible, justified by a simple counterexample. 

The dichotomous modeling of similarity is a major drawback of the binary relation based view: two elements are viewed as being either similar or not similar without anything in-between \cite{Yao00}. In the quantitative setting, one possible solution to this problem is the use of a quantitative similarity measure or a fuzzy similarity relation. Since we are interested in qualitative rather than quantitative similarity here, in this paper we take a different approach by studying fragments of the general framework. More formally, in \prettyref{§:kl} we introduce the $(k,\ell)$-fragments consisting only of generalizations containing at most $\ell$ occurrences of $k$ variables giving raise to a parameterized similarity relation, where we briefly illustrate the concept by exploring the monolinear fragment $(1,1)$ (consisting only of generalizations with exactly one occurrence of a single variable) in the domain of sets and numbers.

In \prettyref{§:Restricted}, we study similarity in three different restricted classes of algebras. In \prettyref{§:MUA}, we consider monounary algebras consisting of a single unary function, where we show in Theorems \ref{t:a_not_approx_b} and \ref{t:a_not_approx_a_theta} that similarity is in general not compatible with congruences. In \prettyref{§:FUA}, we look at finite unary algebras where similarity is closely related to finite automata and regular languages which yields algorithms for its computation. Finally, in \prettyref{§:FA} we establish a tight connection between similarity in finite algebras and tree automata by showing that the latter can be used to compute $k$-similarity.

The idea of this paper to use generalizations to define similarity in the algebraic setting of universal algebra appears to be original. As we construct similarity from first principles using only elementary concepts of universal algebra, to convince the reader of its plausibility we need to validate it either empirically or --- what we prefer here --- theoretically by showing that it fits naturally into the overall mathematical landscape. For this, we show in \prettyref{§:Math} that three different fundamental relations occurring in mathematics are instances of similarity, namely modular arithmetic (\prettyref{§:Modular}), Green's relations \cite{Green51} in semigroups (\prettyref{§:Semigroups}), and the conjugacy relation in group theory where we show that every group is similar to its factor groups (\prettyref{§:Groups}). This shows that the motivation for formulating and studying similarity on an abstract level is not unsubstantiated. Finally, in \prettyref{§:Logical_Interpretation} we show that the purely algebraic notion of similarity can be naturally embedded into first-order logic via model-theoretic types. More precisely, we show that sets of generalizations are in one-to-one correspondence with so-called \textit{g-formulas} and \textit{g-types}, which is appealing as types play a fundamental role in model theory and showing that our notion --- which is primarily motivated by simple examples --- has a natural logical interpretation, provides strong evidence for its suitability.

Finally, in \prettyref{§:TCS_and_AI} we sketch some potential applications to theoretical computer science and artificial intelligence.

We conclude with \prettyref{§:Conclusion} where we give an outlook for future research.


\section{Preliminaries}

We assume the reader to be fluent in basic universal algebra as it is presented for example in \cite[§II]{Burris00}.

Let $\mathbb N_k:=\{k,k+1,\ldots\}$ be the natural numbers starting at some $k\geq 0$. For a natural number $m\in \mathbb N_1$, define $\langle m\rangle := \{1,\ldots,m\}$.

A \textit{\textbf{language}} $L$ of algebras is a set of \textit{\textbf{function symbols}} together with a \textit{\textbf{rank function}} $r:L\to\mathbb N_0$, and a denumerable set $X$ of \textit{\textbf{variables}} distinct from $L$. We omit \textit{\textbf{constant symbols}} as we identify them with 0-ary function symbols. Terms are formed as usual from variables in $X$ and function symbols in $L$ and we denote the set of all such $L$-terms by $T_{L,X}$. The \textit{\textbf{rank}} of a term $s$ is given by the number of variables occurring in $s$ and is denoted by $r(s)$.

An \textit{\textbf{$L$-algebra}} $\mathfrak A$ consists of a non-empty set $\mathbb A$, the \textit{\textbf{universe}} of $\mathfrak A$, and for each function symbol $f\in L$, a function $f^\mathfrak A: \mathbb A^{r(f)}\to \mathbb A$, the \textit{\textbf{functions}} of $\mathfrak A$ (the \textit{\textbf{distinguished elements}} of $\mathfrak A$ are the 0-ary functions). Every term $s$ induces a function $s^\mathfrak A$ on $\mathfrak A$ in the usual way.

A \textit{\textbf{homomorphism}} is a mapping $\bullet: \mathfrak{A\to B}$ between two $L$-algebras satisfying, for each function symbol $f\in L$ and elements $a_1,\ldots,a_{r(f)}\in \mathbb A$,
\begin{align*} 
	f^ \mathfrak A(a_1,\ldots, a_{r(f)})^\bullet = f^ \mathfrak B(a_1^\bullet,\ldots,a_{r(f)}^\bullet).
\end{align*} An \textit{\textbf{isomorphism}} is a bijective homomorphism.

\section{Generalization-based similarity}

In this section, we introduce an abstract algebraic and \textit{qualitative} notion of similarity based on the observation that sets of generalizations contain important information about elements (see the discussion in \prettyref{§:I}).

Let $\mathfrak A = ( \mathbb A,L^ \mathfrak A)$ and $\mathfrak B = ( \mathbb B, L^ \mathfrak B)$ be $L$-algebras over some joint (ranked) language of algebras $L$. We will always write $\mathfrak A$ instead of $\mathfrak{(A,A)}$.

\begin{definition} Define the \textit{\textbf{set of generalizations}} of an element $a\in \mathbb A$ by\footnote{By ``$T_{L,X}\setminus\{a\}$'' we mean the set of $L$-terms different from the constant symbol ``$a$'' in case it is included in the language $L$.}
\begin{align*} 
	\uparrow_ \mathfrak A a := \left\{s\in T_{L,X}\setminus\{a\} \;\middle|\; a = s^\mathfrak A( \textbf{o}),\text{ for some $\textbf{o}\in \mathbb A^{r(s)}$}\right\},
\end{align*} extended to elements $a\in \mathbb A$ and $b\in \mathbb B$ by
\begin{align*} 
	a\uparrow_ \mathfrak{(A,B)} b := (\uparrow_ \mathfrak A a)\cap (\uparrow_\mathfrak B b).
\end{align*} In case $s\in\ \uparrow_ \mathfrak A a$, we say that $s$ \textit{\textbf{generalizes}} $a$ in $\mathfrak A$. A generalization is \textit{\textbf{trivial}} in $\mathfrak{(A,B)}$ iff it generalizes all elements in $\mathbb A$ and $\mathbb B$ and we denote the set of all such trivial generalizations by $\{\;\}_ \mathfrak{(A,B)}$. We will sometimes omit the reference to the underlying algebras in case they are known from the context.

\todo[inline]{sollten wir nicht alle terme $s$ ausschließen, die zur konstante $a$ ``aequivalent'' sind im sinne von $\downarrow s = \{a\}$?!}
\end{definition}


\begin{fact} The generalization operator $\uparrow$ is symmetric in the sense that
\begin{align*} 
	a\uparrow_ \mathfrak{(A,B)} b = b\uparrow_{\mathfrak{(B,A)}} a.
\end{align*} 
\end{fact}

\begin{fact} If $\mathbb A$ is finite and $s$ is injective on $\mathbb A$, then $s$ is a trivial generalization on $\mathfrak A$.
\end{fact}
\begin{proof} A direct consequence of the fact that injective functions on finite sets are bijective.
\end{proof}

\begin{example} In the algebra $\mathfrak M = (\mathbb N_2,\cdot,\mathbb N_2)$ of the natural numbers starting at 2 together with multiplication,\footnote{The second $\mathbb N$ in $(\mathbb N,\cdot,\mathbb N)$ indicates that  \textit{every} natural number is a \textit{distinguished} element and can thus be used as a constant (symbol) to form terms and generalizations.} evenness is captured via
\begin{align*} 
	a\in \mathbb N_2\text{ is even } \quad\Leftrightarrow\quad 2x\in\ \uparrow_ \mathfrak M a,
\end{align*} and primality via
\begin{align*} 
	a\in \mathbb N_2\text{ is prime } \quad\Leftrightarrow\quad \uparrow_ \mathfrak M a = \{x,a\}.
\end{align*}
\end{example}
 
The above example shows how generalizations encode \textit{properties} of the generalized elements. We can thus interpret the set of all generalizations of an element as the set of all properties expressible within the surrounding algebra --- the set of shared properties of two elements $a\in \mathbb A$ and $b\in \mathbb B$ is therefore captured by $a\uparrow_ \mathfrak{(A,B)} b$.

The above discussions motivate the following algebraic definition of similarity via generalizations:

\begin{definition}\label{d:approx} We define the \textit{\textbf{similarity relation}} as follows:
\begin{enumerate}
	\item We say that $a\lesssim b$ \textit{\textbf{holds}} in $\mathfrak{(A,B)}$ --- in symbols,
	\begin{align*} 
		\mathfrak{(A,B)} \models a\lesssim b,
	\end{align*} iff
	\begin{enumerate}
		\item either $(\uparrow_ \mathfrak A a)\cup (\uparrow_ \mathfrak B b)$ consists only of trivial generalizations; or
		\item $a\uparrow_ \mathfrak{(A,B)} b$ contains at least one non-trivial generalization and is maximal with respect to subset inclusion among the sets $a\uparrow_ \mathfrak{(A,B)} c$, $c\in \mathbb B$. That is, for any element $c\in \mathbb B$,\footnote{Note that the first inclusion is a strict inclusion!}
		\begin{align*} 
			\{\;\}_ \mathfrak{(A,B)}\subsetneq a\uparrow_ \mathfrak{(A,B)} b\subseteq a\uparrow_ \mathfrak{(A,B)} c \quad\Rightarrow\quad a\uparrow_ \mathfrak{(A,B)} c\subseteq a\uparrow_ \mathfrak{(A,B)} b.
		\end{align*} We abbreviate the above requirement by simply saying that $a\uparrow_ \mathfrak{(A,B)} b$ is \textit{\textbf{$b$-maximal}}.
	\end{enumerate}

	\item Finally, the \textit{\textbf{similarity relation}} is defined as
	\begin{align*} 
		\mathfrak{(A,B)} \models a\approx b 
			\quad:\Leftrightarrow\quad \mathfrak{(A,B)} \models a\lesssim b \quad\text{and}\quad \mathfrak{(B,A)} \models b\lesssim a,
	\end{align*} in which case we say that $a$ and $b$ are \textit{\textbf{similar}} in $\mathfrak{(A,B)}$.
\end{enumerate}
\end{definition}


\begin{fact} We have
\begin{align}\label{eq:subseteq} 
	\uparrow_ \mathfrak A a\subseteq\ \uparrow_\mathfrak B b \quad\Rightarrow\quad \mathfrak{(A,B)} \models a\lesssim b.
\end{align} and 
\begin{align}\label{eq:=} 
	\uparrow_ \mathfrak A a = \ \uparrow_\mathfrak B b \quad\Rightarrow\quad \mathfrak{(A,B)} \models a\approx b.
\end{align}
\end{fact}

We demonstrate similarity by giving some illustrative examples:


\begin{example} In any algebra $(\mathbb A)$ containing no functions, all elements are similar as expected.
\end{example}

\begin{example}\label{e:a_b_ast} In the algebra $(\{a,b,\ast\},f)$ given by
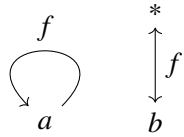
\begin{center}
\begin{tikzpicture} 
    \node (a)               {$a$};
    \node (b) [right=of a]  {$b$};
    \node (c) [above=of b]  {$\ast$};
    \draw[->] (a) to [edge label'={$f$}] [loop] (a);
    \draw[<->] (b) to [edge label'={$f$}] (c);
\end{tikzpicture}
\end{center} we have
\begin{align*} 
	\uparrow a = \left\{f^n(x) \;\middle|\; n\geq 0\right\}
\end{align*} and
\begin{align*} 
	\uparrow b = \left\{f^n(x) \;\middle|\; b = f^n(b), \text{$n$ is even}\right\}\cup \left\{f^n(x) \;\middle|\; b = f^n(\ast), \text{$n$ is odd}\right\} =\ \uparrow a,
\end{align*} and therefore
\begin{align*} 
	a\approx b.
\end{align*} This means that our framework detects the similarity between the loop at $a$ and the circle going through $b$.
\end{example}

\begin{example} Consider the algebra given by
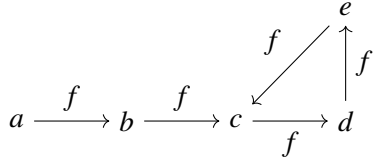
\begin{center}
\begin{tikzpicture} 
	\node (a) {$a$};
	\node (b) [right=of a] {$b$};
	\node (c) [right=of b] {$c$};
	\node (d) [right=of c] {$d$};
	\node (e) [above=of d] {$e$};
	\draw[->] (a) to [edge label={$f$}] (b);
	\draw[->] (b) to [edge label={$f$}] (c);
	\draw[->] (c) to [edge label'={$f$}] (d);
	\draw[->] (d) to [edge label'={$f$}] (e);
	\draw[->] (e) to [edge label'={$f$}] (c);
\end{tikzpicture}
\end{center} We compute
\begin{align*} 
	\uparrow a &= \{x\},\\
	\uparrow b &= \{x,f(x)\},\\
	\uparrow c &= \left\{f^n(x) \;\middle|\; n\geq 0\right\},\\
		&=\ \uparrow d,\\
		&=\ \uparrow e.
\end{align*} This yields with \prettyref{eq:subseteq} and \prettyref{eq:=}\todo{$<$ wird dadurch nicht ganz beschrieben}:
\begin{align*} 
	a<b<c\approx d\approx e.
\end{align*} This coincides with our intuition that $a$ and $b$ differ from $c,d,e$ which all are part of a circle.
\end{example}

\begin{example} Consider the monounary algebra $(\mathbb Z,f)$, where $f: \mathbb Z\to \mathbb Z$ is defined by
\begin{align*} 
	f(0) &:= 0\\
	f(a) &:=
		\begin{cases}
			a-1 & a>0\\
			a+1 & a<0.
		\end{cases}
\end{align*} This can be visualized as
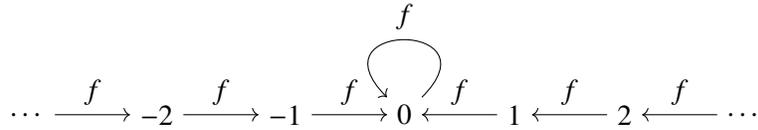
\begin{center}
\begin{tikzpicture} 
	\node (-3) {\ldots};
	\node (-2) [right=of -3] {$-2$};
	\node (-1) [right=of -2] {$-1$};
	\node (0) [right=of -1] {$0$};
	\node (1) [right=of 0] {$1$};
	\node (2) [right=of 1] {$2$};
	\node (3) [right=of 2] {\ldots};
	\draw[->] (-3) to [edge label={$f$}] (-2);
	\draw[->] (-2) to [edge label={$f$}] (-1);
	\draw[->] (-1) to [edge label={$f$}] (0);
	\draw[->] (0) to [edge label'={$f$}][loop] (0);
	\draw[->] (1) to [edge label'={$f$}] (0);
	\draw[->] (2) to [edge label'={$f$}] (1);
	\draw[->] (3) to [edge label'={$f$}] (2);
\end{tikzpicture}
\end{center} We compute
\begin{align*} 
	\uparrow_{(\mathbb Z,f)} a = \left\{f^n(x) \;\middle|\; n\geq 0\right\},\quad\text{for all $a\in\mathbb Z$},
\end{align*} which means that \textit{every} $f^n(x)$ is a trivial generalization and thus
\begin{align*} 
	(\uparrow_{(\mathbb Z,f)} a)\cup (\uparrow_{(\mathbb Z,f)} b) = \{\;\}_{(\mathbb Z,f)}.
\end{align*} Hence,
\begin{align*} 
	(\mathbb Z,f) \models a\approx b,\quad\text{for all $a,b\in\mathbb Z$}.
\end{align*}

On the other hand, in the algebra $(\mathbb N,f)$ of the natural numbers (starting at $0$) with successor
\begin{align*} 
	f(a) := a+1
\end{align*} given by
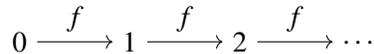
\begin{center}
\begin{tikzpicture} 
	\node (0) [right=of -1] {$0$};
	\node (1) [right=of 0] {$1$};
	\node (2) [right=of 1] {$2$};
	\node (3) [right=of 2] {\ldots};
	\draw[->] (0) to [edge label={$f$}] (1);
	\draw[->] (1) to [edge label={$f$}] (2);
	\draw[->] (2) to [edge label={$f$}] (3);
\end{tikzpicture}
\end{center} we have
\begin{align*} 
	\uparrow_{(\mathbb N,f)}a = \left\{f^n(x) \;\middle|\; 0\leq n\leq a\right\}
\end{align*} which yields
\begin{align*} 
	(\mathbb N,f) \models a\lesssim b \quad\Leftrightarrow\quad a\leq b
\end{align*} and thus
\begin{align*} 
	(\mathbb N,f) \models a\approx b \quad\Leftrightarrow\quad a = b.
\end{align*}
\end{example}

\begin{example}\label{e:48} In the multiplicative arithmetical algebra $\mathfrak M := ( \mathbb N_2,\cdot, \mathbb N_2)$, the following counterexample shows that $\lesssim$ does not capture the divisibility relation as we are going to show
\begin{align}\label{eq:48} 
	4\text{ divides }8 \quad\text{whereas}\quad 4\not\lesssim 8.
\end{align} We have
\begin{align*} 
	\uparrow 4 &= \left\{x,xy,2x,x^2\right\},\\
	\uparrow 8 &= \left\{x,xy,xyz,x^2y,x^3,2x,4x\right\}.
\end{align*} Notice that $x^2$ is a generalization of 4 since it is a square number whereas it is \textit{not} a generalization of 8. Now we compute the set of generalizations of 16, which \textit{is} a square number:
\begin{align*} 
	\uparrow 16 = \left\{x,xy,xyx,wxyx,wxyz,2x,2xy,2xyx,4x,4xy,8x,x^2,x^2y^2,x^2yz,2x^3,x^3y,x^4\right\}.
\end{align*} We see that
\begin{align*} 
	\uparrow 4\subsetneq\ \uparrow 16
\end{align*} and
\begin{align*} 
	4\uparrow 8 = \{x,xy,2x\}\subsetneq \left\{x,xy,2x,x^2\right\} = 4\uparrow 16
\end{align*} and thus \prettyref{eq:48} holds; moreover, it shows
\begin{align*} 
	4\lesssim 16.
\end{align*} We clearly have
\begin{align*} 
	16\not\lesssim 4
\end{align*} which amounts to
\begin{align*} 
	4\lnsim 16.
\end{align*}

Finally, it should be noted that
\begin{align*} 
	\uparrow 4\subsetneq\ \uparrow a
\end{align*} holds for \textit{every} even square number $a$!
\end{example}

\begin{example} Interestingly, in the algebra $\mathfrak M' := (\mathbb N_2,\cdot)$ not containing any distinguished elements, we have for every prime number $p$,
\begin{align*} 
	\uparrow_{ \mathfrak M'} p = \{x\},
\end{align*} that is, $p$ has only the trivial generalization $x$. Hence,
\begin{align}\label{eq:pq} 
	\mathfrak M' \models p\approx q,\quad\text{for any prime numbers $p$ and $q$.}
\end{align}

The situation does not change if we consider the algebra $\mathfrak M$ from \prettyref{e:48} containing distinguished elements for any number in $\mathbb N_2$, even though we now have
\begin{align*} 
	\uparrow_ \mathfrak M p = \{x,p\}
\end{align*} containing the non-trivial generalization $p$. Notice that we again have
\begin{align*} 
	\mathfrak M \models p\approx q,\quad\text{for any prime numbers $p$ and $q$.}
\end{align*}
\end{example}

\begin{example}\label{e:aa} In the word domain $(\{a\}^+,\cdot)$, where $\{a\}^+$ is the set of all non-empty words over $a$ and $\cdot$ denotes the concatenation of words, the following counterexample shows that $\lesssim$ does not coincide with the prefix relation as we are going to prove
\begin{align}\label{eq:aa}
	aa\text{ is a prefix of }aaa \quad\text{whereas}\quad aa\not\lesssim aaa.
\end{align} 
We see that $xx$ is a generalization of $aa$ which is \textit{not} a generalization of $aaa$. Now it is easy to verify that $\uparrow aaaa$ contains all generalizations of $aa$, which directly yields
\begin{align*} 
	aa\uparrow aaa \subsetneq aa\uparrow aaaa,
\end{align*} which proves \prettyref{eq:aa}; moreover, it shows
\begin{align*} 
	aa\lesssim aaaa.
\end{align*}

\end{example}

\section{Properties}

In this section, we study some fundamental properties of similarity.






We start by analyzing the following basic properties below:
\begin{align} 
	\label{eq:r} & \mathfrak A \models a\approx a \quad\text{(reflexivity)},\\
	& \mathfrak{(A,B)} \models a\approx b \quad\Leftrightarrow\quad \mathfrak{(B,A)} \models b\approx a \quad\text{(symmetry)},\\
	\label{eq:t} & \mathfrak{(A,B)} \models a\approx b \quad\text{and}\quad \mathfrak{(B,C)} \models b\approx c \quad\Rightarrow\quad \mathfrak{(A,C)} \models a\approx c \quad\text{(transitivity)}. 
\end{align}

\begin{theorem}\label{t:rst} The similarity relation is reflexive, symmetric, and in general not transitive.
\end{theorem}
\begin{proof} Reflexivity and symmetry hold trivially.


To disprove transitivity, we first consider the three unary algebras
\begin{align*} 
	\mathfrak A = (\{a,a'\},f,g) \quad\text{and}\quad \mathfrak B = (\{b\},f,g) \quad\text{and}\quad \mathfrak C = (\{c,c'\},f,g)
\end{align*} given by
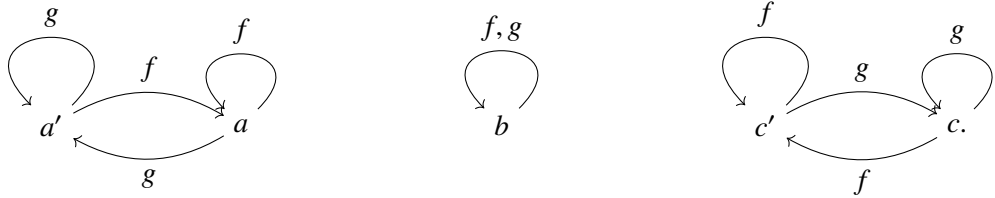
\begin{center}
\begin{tikzpicture}[node distance=2cm and 2cm]
	\node (a') {$a'$};
	\node (a) [right=of a'] {$a$};
	\node (b) [right=of a,xshift=1cm] {$b$};
	\node (c') [right=of b,xshift=1cm] {$c'$};
	\node (c) [right=of c'] {$c$.};
	\draw[->] (a') to [edge label'={$g$}][loop] (a');
	\draw[->] (a) to [edge label'={$f$}][loop] (a);
	\draw[->] (a') to [edge label={$f$}][bend left] (a);
	\draw[->] (a) to [edge label={$g$}][bend left] (a');
	\draw[->] (b) to [edge label'={$f,g$}][loop] (b);
	\draw[->] (c') to [edge label'={$f$}][loop] (c');
	\draw[->] (c) to [edge label'={$g$}][loop] (c);
	\draw[->] (c') to [edge label={$g$}][bend left] (c);
	\draw[->] (c) to [edge label={$f$}][bend left] (c');
\end{tikzpicture}
\end{center} Since $b$ is the only element in $\mathfrak B$ and since $a\uparrow_ \mathfrak{(A,B)} b$ contains non-trivial generalizations, we must have
\begin{align*} 
	\mathfrak{(A,B)} \models a\lesssim b.
\end{align*} Moreover,
\begin{align*} 
	g(f(x))\in b\uparrow_\mathfrak{(B,C)} c \quad\text{whereas}\quad g(f(x))\not\in b\uparrow_\mathfrak{(B,C)} c'
\end{align*} shows the $c$-maximality and non-emptiness (i.e. there are non-trivial generalizations) of $b\uparrow_\mathfrak{(B,C)} c$ which implies
\begin{align*} 
	\mathfrak{(B,C)} \models b\lesssim c.
\end{align*} 

On the other hand,
\begin{align*} 
	a\uparrow_\mathfrak{(A,C)} c'=\ \uparrow_ \mathfrak A a
\end{align*} and
\begin{align*} 
	f(g(x))\in a\uparrow_\mathfrak{(A,C)} c' \quad\text{whereas}\quad f(g(x))\not\in a\uparrow_\mathfrak{(A,C)}c
\end{align*} shows
\begin{align*} 
	a\uparrow_\mathfrak{(A,C)}c\subsetneq a\uparrow_\mathfrak{(A,C)} c'
\end{align*} which implies
\begin{align*} 
	\mathfrak{(A,C)} \models a\not\lesssim c.
\end{align*}

The proof can easily be adapted to the single algebra consisting of the three algebras $\mathfrak{A,B,C}$ which means that transitivity may fail even in a single algebra.
\todo[inline]{Der Beweis ist noch nicht ganz fertig, da ich nur $\lesssim$ aber nicht $\approx$ behandelt habe --- er gilt aber offensichtlich auch fuer $\approx$}
\end{proof}



\subsection{Homomorphisms}

It is reasonable to expect isomorphisms --- which are bijective structure-preserving mappings between algebras --- to be compatible with similarity.

\begin{example} Let $A := \{a\}$ be the alphabet consisting of the single letter $a$, and let $A^\ast$ denote the set of all words over $A$ including the empty word $\varepsilon$. We can identify every sequence $a^n=a\ldots a$ ($n$ consecutive $a$'s) with the non-negative integer $n$, for every $n\geq 0$. Therefore, define the isomorphism $\bullet:(\mathbb N,+)\to (A^\ast,\cdot)$ via
\begin{align*} 
    0^\bullet := \varepsilon \quad\text{and}\quad n^\bullet := a^n,\quad n\geq 1.
\end{align*} We expect
\begin{align*} 
    ((\mathbb N,+),(A^\ast,\cdot)) \models n\approx n^\bullet,\quad\text{for all $n\geq 0$.}
\end{align*} That this is indeed the case is the content of the First Homomorphism \prettyref{t:FIT} below.
\end{example}

\begin{lemma}[Homomorphism Lemma]\label{l:HL} For any homomorphism $\bullet:\mathfrak{A\to B}$ and element $a\in \mathbb A$, we have
\begin{align*} 
	\uparrow_ \mathfrak A a\subseteq \ \uparrow_\mathfrak B a^\bullet.
\end{align*} In case $\bullet$ is an isomorphism, we have
\begin{align*} 
	\uparrow_ \mathfrak A a = \ \uparrow_\mathfrak B a^\bullet.
\end{align*} 
\end{lemma}
\begin{proof} Since $\bullet$ is a homomorphism, we have
\begin{align*} 
	s\in\ \uparrow_ \mathfrak A a 
		\quad&\Leftrightarrow\quad a = s^\mathfrak A( \textbf{o}),\quad\text{for some $\textbf{o}\in \mathbb A^{r(s)}$},\\
		&\Rightarrow\quad a^\bullet = s^\mathfrak A(\textbf{o})^\bullet = s^ \mathfrak B(\textbf{o}^\bullet),\\
		&\Rightarrow\quad s\in\ \uparrow_\mathfrak B a^\bullet.
\end{align*} 

In case $\bullet$ is an isomorphism, $\bullet^{-1}$ is an isomorphism as well, and thus we further have
\begin{align*} 
	s\in\ \uparrow_\mathfrak B a^\bullet
		\quad&\Leftrightarrow\quad a^\bullet = s^\mathfrak B(\textbf{o}),\quad\text{for some $\textbf{o}\in \mathbb B^{r(s)}$},\\
		&\Rightarrow\quad a = s^\mathfrak B(\textbf{o})^{\bullet^{-1}} = s^\mathfrak A(\textbf{o}^{\bullet^{-1}})\\
		&\Rightarrow\quad s\in\ \uparrow_ \mathfrak A a.
\end{align*}
\end{proof}

\begin{theorem}[First Isomorphism Theorem]\label{t:FIT} For any homomorphism $\bullet: \mathfrak{A\to B}$ and element $a\in \mathbb A$,
\begin{align*} 
	\mathfrak{(A,B)} \models a\lesssim a^\bullet.
\end{align*} In case $\bullet$ is an isomorphism, we have
\begin{align*} 
	\mathfrak{(A,B)} \models a\approx a^\bullet.
\end{align*}
\end{theorem}
\begin{proof} An immediate consequence of the Homomorphism \prettyref{l:HL} and \prettyref{eq:=}.
\end{proof}

The next two examples are simple yet important as they show that similarity cannot be captured by homomorphisms (which thus formalize a different notion of structural similarity):

\begin{example}\label{e:H_1} Let the monounary algebras $\mathfrak A := (\{a,b\},f)$ and $\mathfrak B := (\{c\},f)$ be given by
\begin{center}
\begin{tikzpicture}[node distance=2cm and 2cm]
	\node (b) {$b$};
	\node (a) [below=of b] {$a$};
	\node (c) [right=of a] {$a^\bullet$};
	\draw[->] (a) to [edge label={$f$}] (b); 
	\draw[->] (b) to [edge label'={$f$}] [loop] (b);
	\draw[->] (c) to [edge label'={$f$}] [loop] (c);
	\draw[dashed,->] (a) to [edge label'={$\bullet$}] (c);
	\draw[dashed,->] (b) to [edge label={$\bullet$}] (c);
\end{tikzpicture}
\end{center} The mapping $\bullet$ as depicted above is obviously a homomorphism. However, we have
\begin{align*} 
	a^\bullet\uparrow_{\mathfrak{(B,A)}} a &= \{x\}\subsetneq \left\{f^n(x) \;\middle|\; n\geq 0\right\} = a^\bullet\uparrow_{\mathfrak{(B,A)}} b
\end{align*} which shows
\begin{align*} 
	\mathfrak{(B,A)} \models a^\bullet\not\lesssim a
\end{align*} and thus
\begin{align*} 
	\mathfrak{(A,B)} \models a\not\approx a^\bullet.
\end{align*}
\end{example}



\begin{example} Let the monounary algebras $\mathfrak A := (\{a\},f)$ and $\mathfrak B := (\{b,c\},f)$ be given by
\begin{center}
\begin{tikzpicture}[node distance=2cm and 2cm]
	\node (a) {$a$};
	\node (b) [right=of a] {$b$};
	\node (c) [below=of b] {$c$};
	\draw[->] (a) to [edge label'={$f$}] [loop] (a);
	\draw[-,dashed] (a) to [edge label={$\approx$}] (b);
	\draw[<->] (b) to [edge label={$f$}] (c); 
\end{tikzpicture}
\end{center} We then have 
\begin{align*} 
	\mathfrak{(A,B)} \models a\approx b,
\end{align*} which is shown as in \prettyref{e:a_b_ast}. However, $a^\bullet := b$ is not a homomorphism since
\begin{align*} 
	f(a)^\bullet = a^\bullet = b \quad\text{whereas}\quad f(a^\bullet) = f(b) = c.
\end{align*}
\end{example}

\begin{theorem}[Second Isomorphism Theorem]\label{t:SIT} For any isomorphism $\bullet:\mathfrak{A\to B}$ and elements $a,b\in \mathbb A$,
\begin{align*} 
	\mathfrak A \models a\approx b \quad\Leftrightarrow\quad \mathfrak B \models a^\bullet\approx b^\bullet.
\end{align*}
\end{theorem}
\begin{proof} An immediate consequence of the Homomorphism \prettyref{l:HL}.
\end{proof}

\section{\texorpdfstring{The $(k,\ell)$-fragments}{Fragments}}\label{§:kl}

In this section, we introduce the $(k,\ell)$-fragments of similarity consisting only of generalizations containing at most $\ell$ occurrences of $k$ variables giving raise to a parameterized similarity relation:

\begin{definition}\label{d:k_ell} Let $X_k := \{x_1,\ldots,x_k\}$, for $k\in\mathbb N\cup\{\infty\}$, so that $X_\infty = X$. Now for $k,\ell\in \mathbb N\cup\{\infty\}$, define
\begin{align*} 
	\uparrow^{(k,\ell)}_ \mathfrak A a := (\uparrow_ \mathfrak A a)\cap \left\{s(x_1,\ldots,x_k)\in T_{L,X_k} \;\middle|\; \text{each of the $k$ variables in $X_k$ occurs at most $\ell$ times in $s$}\right\}.
\end{align*} We write $k$ instead of $(k,\infty)$ so that
\begin{align*} 
	\uparrow^k_ \mathfrak A a=(\uparrow_ \mathfrak A a)\cap T_{L,X_k}.
\end{align*} Every $(k,\ell)$-fragment gives raise to a similarity relation $\approx_{(k,\ell)}$ defined as $\approx$ with $X$ replaced by $X_k$ and $\uparrow$ replaced by $\uparrow^{(k,\ell)}$ (see \prettyref{d:approx}). The simplest fragment --- namely, the $(1,1)$-fragment --- consists of \textit{\textbf{monolinear generalizations}} containing exactly one occurrence of a single variable $x$. We denote the \textit{\textbf{set of monolinear generalizations}} of $a$ in $\mathfrak A$ by $\uparrow^m_ \mathfrak A a$ and the so-obtained similarity relation by $\approx_m$. Moreover, we consider the \textit{\textbf{linear}} $(\infty,1)$-fragment consisting only of generalizations with at most one occurrence of each variable, and we denote the \textit{\textbf{set of linear generalizations}} of $a$ in $\mathfrak A$ by $\uparrow^l_ \mathfrak A a$.
\end{definition}

For the reader's convenience, we shall now illustrate some simple fragments. The following subsections are technically trivial and serve only the purpose of showing that even in the most simple fragment similarity may have a non-trivial meaning.

\subsection{Monolinear set similarity}

In the domain of sets, we obtain the following characterization of similarity in the monolinear fragment. Let $U$ be a set. In $(2^U,.^c)$, where $.^c$ denotes the set complement operation, we have for every $A\subseteq U$ (here $X$ denotes a variable standing for a set and not the set of variables from above)
\begin{align*} 
	\uparrow^m A = \{X,X^c\}.
\end{align*} Hence, we have the trivial characterization of monolinear similarity given by
\begin{align*} 
	A\approx_m B,\quad\text{for all $A,B\subseteq U$.}
\end{align*}

\subsection{\texorpdfstring{Additive natural number $(1,\infty)$-similarity}{Additive natural number similarity}}

In $(\mathbb N,+)$ we have
\begin{align*} 
	\uparrow^{(1,\infty)} a
		&=\{kx\mid a=ko,\text{ for some }o\in\mathbb N\}\\
		&=\{kx\mid\text{$k$ divides $a$}\},
\end{align*} where $kx$ is an abbreviation for the term $x+\ldots+x$ ($k$ times) not containing constants. Hence,
\begin{align*} 
	a\uparrow^{(1,\infty)} b=\{kx\mid k\text{ divides $a$ and $b$}\},
\end{align*} which is $b$-maximal iff $a$ divides $b$, that is,
\begin{align*} 
	a\lesssim_{(1,\infty)} b \quad\Leftrightarrow\quad \text{$a$ divides $b$}.
\end{align*} That is, within the monolinear fragment the relation $\lesssim_{(1,\infty)}$ captures divisibility of natural numbers in the algebra $(\mathbb N,+)$. This implies
\begin{align}\label{eq:1_infty} 
	a\approx_{(1,\infty)} b \quad\Leftrightarrow\quad \text{$a$ divides $b$ and $b$ divides $a$} \quad\Leftrightarrow\quad a=b.
\end{align}

\subsection{Linear additive natural number similarity}

In $(\mathbb N_1,+)$, we have
\begin{align*} 
	\uparrow^\ell a=\{x_1+\ldots +x_n\mid 1\leq n\leq a\},
\end{align*} which means that $a\uparrow^\ell b$ is $b$-maximal iff $a\leq b$. Hence,
\begin{align*} 
	a\approx_\ell b \quad\Leftrightarrow\quad a=b.
\end{align*} Notice that by \prettyref{eq:1_infty}, we have (recall that the linear fragment corresponds to the $(\infty,1)$-fragment)
\begin{align*} 
	a\approx_{(\infty,1)} b \quad\Leftrightarrow\quad a\approx_{(1,\infty)} b.
\end{align*}

\subsection{Linear word similarity}

In $(A^+,\cdot)$, where $A$ is a finite non-empty alphabet, we have for every word $\textbf{a}\in \mathbb A^+$,\footnote{Here $|\textbf{a}|$ denotes the number of symbols occurring in $\textbf{a}$.}
\begin{align*} 
	\uparrow^\ell \textbf{a}=\left\{x_1\ldots x_n \;\middle|\; \textbf{a}= \textbf{b}_1\ldots \textbf{b}_n, \text{ for some }\textbf{b}_1,\ldots ,\textbf{b}_n\in \mathbb A^+, 1\leq n\leq |\textbf{a}|\right\}.
\end{align*} Hence, $\textbf{a}\uparrow^\ell \textbf{b}$ is $\textbf{b}$-maximal iff $|\textbf{a}|\leq |\textbf{b}|$, which implies
\begin{align*} 
	\textbf{a}\approx_\ell \textbf{b} \quad\Leftrightarrow\quad |\textbf{a}|=|\textbf{b}|.
\end{align*}

\section{Restricted algebras}\label{§:Restricted}

In this section, we study similarity in three restricted classes of algebras, namely monounary \prettyref{§:MUA}, finite unary \prettyref{§:FUA}, and finite algebras \prettyref{§:FA}.

\subsection{Monounary algebras}\label{§:MUA}

In this section, we study similarity in monounary algebras consisting of a single unary function, where we show that similarity and congruences are, in general, incompatible concepts.

Recall that an equivalence relation $\theta$ on a monounary algebra $\mathfrak A=(\mathbb A,S)$ is a \textit{\textbf{congruence}} iff
\begin{align*} 
	a\theta b \quad\Rightarrow\quad S(a)\theta S(b),\quad\text{for all $a,b\in \mathbb A$.}
\end{align*} The \textit{\textbf{factor algebra}} obtained from $\mathfrak A$ with respect to $\theta$ is given by
\begin{align*} 
	\mathfrak A/\theta := (\mathbb A/\theta,S),
\end{align*} where
\begin{align*} 
	\mathbb A/\theta := \{[a]_\theta\mid a\in \mathbb A\}
\end{align*} contains the congruence classes
\begin{align*} 
	[a]_\theta := \{b\in \mathbb A\mid a\theta b\}
\end{align*} with respect to $\theta$, and $S:\mathbb A/\theta\to \mathbb A/\theta$ is defined by
\begin{align*} 
	S([a]_\theta) := [S(a)]_\theta,\quad\text{for all $[a]_\theta\in A/\theta$}.
\end{align*}

\begin{theorem}\label{t:a_not_approx_b} There is a finite monounary algebra $\mathfrak A = (\langle m\rangle,S)$, a congruence $\theta$, and elements $a,b\in \langle m\rangle$ such that
\begin{align*} 
	\mathfrak A \models a \not\approx b \quad\text{whereas}\quad \mathfrak A/\theta \models [a]_\theta \approx [b]_\theta.
\end{align*}
\end{theorem}
\begin{proof} Define $\mathfrak A$ by
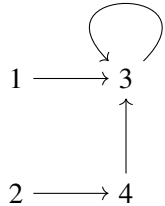
\begin{center}
\begin{tikzpicture} 
	\node (1) {1};
	\node (2) [below=of 1] {2};
	\node (3) [right=of 1] {3};
	\node (4) [right=of 2] {4};
	\draw[->] (1) to (3);
	\draw[->] (2) to (4);
	\draw[->] (3) to [loop] (3);
	\draw[->] (4) to (3);
\end{tikzpicture}
\end{center} and define the congruence
\begin{align*} 
	\theta := \{\{1,2\},\{3,4\}\}
\end{align*} giving raise to the factor algebra
\begin{center}
\begin{tikzpicture} 
	\node (1) {$[1]_\theta$};
	\node (2) [right=of 1] {$[3]_\theta$};
	\draw[->] (1) to (2);
	\draw[->] (2) to [loop] (2);
\end{tikzpicture}
\end{center} 

Since
\begin{align*} 
	\uparrow_ \mathfrak A 3= \mathbb N \quad\text{whereas}\quad \uparrow_ \mathfrak A 4=\{0,1\},
\end{align*} we have
\begin{align*} 
	\mathfrak A \models 3\not\approx 4.
\end{align*} On the other hand, sine $[3]_\theta = [4]_\theta$ we have
\begin{align*} 
	\mathfrak A/\theta \models [3]_\theta\approx [4]_\theta
\end{align*} by the reflexivity of similarity.
\end{proof}

\begin{theorem}\label{t:a_not_approx_a_theta} There is a finite monounary algebra $\mathfrak A = (\langle m\rangle,S)$, a congruence $\theta$, and elements $a,b\in \langle m\rangle$ such that
\begin{align*} 
	(\mathfrak A, \mathfrak A/\theta) \models a \not\approx [a]_\theta.
\end{align*}
\end{theorem}
\begin{proof} Reconsider the finite monounary algebra $\mathfrak A$ and congruence $\theta$ in the proof of \prettyref{t:a_not_approx_b}. Since
\begin{align*} 
	\uparrow_ \mathfrak A 3= \mathbb N,\quad \uparrow_ \mathfrak A 4=\{0,1\} \quad\text{whereas}\quad \uparrow_{ \mathfrak A/\theta} [4]_\theta=\ \uparrow_{ \mathfrak A/\theta} [3]_\theta= \mathbb N,
\end{align*} we have
\begin{align*} 
	& ( \mathfrak A, \mathfrak A/\theta) \models 3\approx [4]_\theta\\
	& (\mathfrak A, \mathfrak A/\theta) \models 4\lnsim [4]_\theta
\end{align*} and thus
\begin{align*} 
	(\mathfrak A, \mathfrak A/\theta) \models 4\not\approx [4]_\theta.
\end{align*}
\end{proof}

\subsection{Finite unary algebras a.k.a semiautomata}\label{§:FUA}

In this section, we study generalization-based similarity in finite unary algebras, which can be seen as semiautomata, and show that we can use well-known methods from the theory of finite automata.

In the rest of this section, let
\begin{align*} 
	\mathfrak A = (\mathbb A,\Sigma := \{f_1,\ldots,f_n\}),
\end{align*} for some $n\geq 1$, be a finite unary algebra with finite universe $\mathbb A$. We shall now recall that every such algebra is essentially a semiautomaton.

Recall that a (\textit{\textbf{finite deterministic}}) \textit{\textbf{semiautomaton}} (see e.g. \cite[§2.1]{Holcombe82}) is a construct $$\mathfrak S = (S,\Sigma,\delta),$$ where $S$ is a finite set of \textit{\textbf{states}}, $\Sigma$ is a finite \textit{\textbf{input alphabet}}, and $\delta : S\times \Sigma\to S$ is a \textit{\textbf{transition function}}. Every semiautomaton can be seen as a finite unary algebra in the following well-known way: every symbol $\sigma\in\Sigma$ induces a unary function $\sigma^ \mathfrak S : S\to S$ via $\sigma^ \mathfrak S := \delta(x,\sigma)$. We can now omit $\delta$ and define
\begin{align*} 
	\mathfrak S' := (S,\Sigma^ \mathfrak S:=\{\sigma^ \mathfrak S\mid \sigma\in\Sigma\}).
\end{align*} It is immediate from the construction that $\mathfrak S$ and $\mathfrak S'$ represent essentially the same semiautomaton and that every semiautomaton can be represented in that way --- the difference is that $\mathfrak S'$ is a finite unary algebra!

Recall that a (\textit{\textbf{finite deterministic}}) \textit{\textbf{automaton}} (see e.g. \cite[§1.1]{Sipser13}) is a construct $$\mathscr A := (Q,\Sigma,\delta,q_0,F),$$ where $(Q,\Sigma,\delta)$ is a semiautomaton, $q_0\in Q$ is the \textit{\textbf{initial state}}, and $F\subseteq Q$ is a set of \textit{\textbf{final states}}. 

The \textit{\textbf{behavior}} of $\mathscr A$ is given by
\begin{align*} 
	|| \mathscr A|| := \left\{w\in \Sigma^\ast \;\middle|\; \delta^\ast(q_0,w)\in F \right\},
\end{align*} where $\Sigma^\ast$ denotes the set of all words over $\Sigma$ including the \textit{\textbf{empty word}} $\varepsilon$ and $\delta^\ast : Q\times\Sigma^\ast\to Q$ is defined recursively as follows, for $q\in Q,a\in\Sigma,w\in\Sigma^\ast$:
\begin{align*} 
	\delta^\ast(q,\varepsilon) &:= q,\\
	\delta^\ast(q,aw) &:= \delta^\ast(\delta(q,a),w).
\end{align*} 

Notice that since automata are built from semiautomata by adding an initial state and a set of final states, and since every semiautomaton $\mathfrak S = (S,\Sigma,\delta)$ can be represented in the form of a finite unary algebra $\mathfrak S'=(S,\Sigma^ \mathfrak S)$ as above, we can reformulate every automaton $\mathscr A=(Q,\Sigma,\delta,q_0,F)$ as $\mathscr A' = (Q,\Sigma^ \mathscr A,q_0,F)$, where $\Sigma^ \mathscr A := \{\sigma^ \mathscr A\mid \sigma\in\Sigma\}$ and $\sigma^ \mathscr A := \delta(\,.\,,\sigma):Q\to Q$. In other words, given a finite unary algebra (semiautomaton) $$\mathfrak A = (\mathbb A,\Sigma),$$ we can construct a finite automaton
\begin{align*} 
	\mathfrak A_{a\to F} = (\mathbb A,\Sigma,a,F)
\end{align*} by designating a state $a\in \mathbb A$ as the initial state, and by designating a set of states $F\subseteq \mathbb A$ as final states.

We want to compute the set of generalizations $\uparrow_ \mathfrak A a$. Notice that we can identify, for example, each term in $T_{\{f,g\}}(\{x\})$ with a word over the alphabet $\Sigma=\{f,g\}$: for instance, the term $fgfx$ can be identified with the word $fgf\in\Sigma^\ast$ since the variable $x$ contains no information. We denote the function induced by a word $w\in\Sigma^\ast$ in $\mathfrak A$ by $w^ \mathfrak A$ --- for example, $(fg)^ \mathfrak A$ is the function on $A$ which first applies $g$ and then $f$. 

We shall now show that in any finite unary algebra (semiautomaton) $\mathfrak A = (\mathbb A,\Sigma)$, the set of generalizations $\uparrow_ \mathfrak A a$ can be computed by some finite automaton as illustrated by the following example: 

\begin{example}\label{e:A} Consider the finite unary algebra (semiautomaton)
\begin{align*} 
	\mathfrak A = (\{a,b\},\Sigma := \{f,g\})
\end{align*} given by
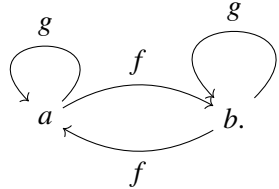
\begin{center}
\begin{tikzpicture}[node distance=2cm and 2cm] 
	\node (a) {$a$};
	\node (b) [right=of a] {$b$.};
	\draw[->] (a) to [edge label={$f$}] [bend left] (b);
	\draw[->] (b) to [edge label={$f$}] [bend left] (a);
	\draw[->] (a) to [edge label'={$g$}] [loop] (a);
	\draw[->] (b) to [edge label'={$g$}] [loop] (b);
\end{tikzpicture}
\end{center} 

We can identify the set of all generalizations of $a$ in $\mathfrak A$ with
\begin{align*} 
	\uparrow_ \mathfrak A a = \left\{w\in\Sigma^\ast \;\middle|\; \delta^\ast(a,w) = a\right\}\cup \left\{u\in\Sigma^\ast \;\middle|\; \delta^\ast(b,u) = a\right\}.
\end{align*} 

Now define the automaton $\mathfrak A_{a\to \{a\}}$ by adding to the semiautomaton $\mathfrak A$ the initial state $a$ and the set of final states $\{a\}$ (we use here the standard pictorial notation for automata) by
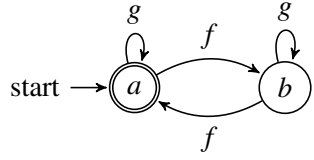
\begin{center}
\begin{tikzpicture}[->,>=stealth',semithick,shorten >= 1pt,node distance=2cm,auto]
	\node[state,initial,accepting] (a) {$a$};
	\node[state] (b) [right of=a] {$b$};
	\path[->]   
	    (a) edge [loop above] node {$g$} (a)
	    (a) edge [bend left] node {$f$} (b)
	    (b) edge [bend left] node {$f$} (a)
	    (b) edge [loop above] node {$g$} (b)
	;
\end{tikzpicture}
\end{center} and the automaton $\mathfrak A_{b\to \{a\}}$ by
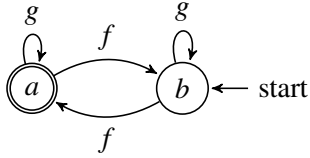
\begin{center}
\begin{tikzpicture}[->,>=stealth',semithick,shorten >= 1pt,node distance=2cm,auto,initial where=right]
	\node[state,accepting] (a) {$a$};
	\node[state,initial] (b) [right of=a] {$b$};
	\path[->]   
	    (a) edge [loop above] node {$g$} (a)
	    (a) edge [bend left] node {$f$} (b)
	    (b) edge [bend left] node {$f$} (a)
	    (b) edge [loop above] node {$g$} (b)
	;
\end{tikzpicture}
\end{center} 

We then clearly have
\begin{align}\label{eq:uparrow_a-b} 
	\uparrow_ \mathfrak A a = ||\mathfrak A_{a\to \{a\}}||\cup ||\mathfrak A_{b\to \{a\}}||.
\end{align}
\end{example}

It is straightforward to generalize the construction in \prettyref{e:A}: 

\begin{definition} Given a finite unary algebra (semiautomaton) $\mathfrak A=(\mathbb A,\Sigma)$, the automaton $\mathfrak A_{b\to \{a\}}$ is the automaton induced by the functions in $\Sigma$ with start state $b$ and single final state $a$ given by
\begin{align*} 
	\mathfrak A_{b\to \{a\}} := (\mathbb A,\Sigma,b,\{a\}).
\end{align*} 
\end{definition}

As a generalization of \prettyref{eq:uparrow_a-b} we obtain:

\begin{fact}\label{f:uparrow_a} Given any finite\footnote{Finiteness is required since regular languages are not closed under \textit{infinite} union.} unary algebra (semiautomaton) $\mathfrak A=(\mathbb A,\Sigma)$ and $a\in \mathbb A$, we have
\begin{align}\label{eq:uparrow_a}
	\uparrow_ \mathfrak A a = \bigcup_{b\in \mathbb A} ||\mathfrak A_{b\to \{a\}}||.
\end{align}
\end{fact}

By \prettyref{f:uparrow_a}, computing
\begin{align*} 
	a\uparrow_ \mathfrak{(A,B)} b = (\uparrow_ \mathfrak A a)\cap (\uparrow_\mathfrak B b)
\end{align*} amounts to computing the intersection of two regular languages by applying standard techniques from automata theory (see e.g. \cite{Sipser13}). Hence, given $a\in \mathbb A$ and $b\in \mathbb B$, we can check $a\stackrel?\lesssim b$ by checking $a\uparrow b\stackrel?\subsetneq a\uparrow c$ for all the \textit{finitely many} $c\neq b\in \mathbb B$ which in total gives us an algorithm for the computation of similarity in finite unary algebras (i.e. semiautomata).


\subsection{Finite algebras}\label{§:FA}

In this section, we study generalization-based similarity in finite algebras and show that it is closely related to regular tree languages and finite tree automata.

Recall that a (\textit{\textbf{frontier-to-root}}) \textit{\textbf{tree automaton}} $\mathfrak T_{\mathfrak A,k,\alpha,F} := (\mathfrak A,L,X_k,\alpha,F)$\footnote{Recall from \prettyref{d:k_ell} that $X_k = \{x_1,\ldots,x_k\}$.}, $k\geq 1$, consists of (see e.g. \cite{Comon08,Gecseg15})
\begin{itemize}
	\item a \textit{finite} $L$-algebra $\mathfrak A$,
	\item an \textit{\textbf{initial assignment}} $\alpha:X_k\to\mathfrak A$, and
	\item a set $F\subseteq A$ of \textit{\textbf{final states}}.
\end{itemize} The \textit{\textbf{regular tree language}} recognized by $\mathfrak T_{\mathfrak A,k,\alpha,F}$ is given by
\begin{align*} 
	||\mathfrak T_{\mathfrak A,k,\alpha,F}|| := \left\{s\in T_{L,X_k} \;\middle|\; s^\mathfrak A\alpha\in F\right\}.
\end{align*} 

Notice that we can write the set of $k$-generalizations of $a$ in $\mathfrak A$ as
\begin{align*} 
	\uparrow^k_ \mathfrak A a = \left\{s\in T_{L,X_k}\setminus\{a\} \;\middle|\; s^\mathfrak A\alpha = a,\text{ for some initial assignment $\alpha$}\right\}.
\end{align*} We thus have
\begin{align*} 
	\uparrow^k_ \mathfrak A a = \bigcup_{\alpha\in \mathbb A^{X_k}}||\mathfrak T_{\mathfrak A,k,\alpha,\{a\}}||.
\end{align*} Since tree automata are closed under finite union and since $\mathbb A^{X_k}$ is finite, there is some tree automaton $\mathfrak T_{\mathfrak A,k,a}$ such that
\begin{align*} 
	\uparrow^k_ \mathfrak A a = ||\mathfrak T_{\mathfrak A,k,a}||.
\end{align*} Moreover, since tree automata are closed under intersection, there is some tree automaton $\mathfrak T_{\mathfrak{(A,B)},k,a,b}$ such that
\begin{align*} 
	a\uparrow^k_ \mathfrak{(A,B)} b = ||\mathfrak T_{\mathfrak{(A,B)},k,a,b}||.
\end{align*} Since the underlying algebras are finite, checking $a$-, and $b$-maximality can be done in time linear to the size of the algebra. In total, we have thus derived an \textbf{algorithm} for the computation of $k$-similarity in finite algebras.
\todo[inline]{Was ist mit $(\uparrow_ \mathfrak A a)\cup (\uparrow_ \mathfrak B b) = \{\;\}$?}

\section{Three fundamental relations in mathematics}\label{§:Math}

To illustrate the expressibility of our notion of similarity, we show that it can model three fundamental relations occurring in mathematics, namely modular arithmetic \prettyref{§:Modular}, Green's relations in semigroups \prettyref{§:Semigroups}, and the conjugacy relation in groups \prettyref{§:Groups}.

\subsection{Modular arithmetic}\label{§:Modular}

Recall that $a\equiv_m b$ (read as ``$a$ is congruent to $b$ mod $m$''; usually written as $a\equiv b \mod m$), for integers $a,b,m$, iff $a=km+r$ and $b=\ell m+r$, for some integers $k,\ell,r$ with $0\leq r<m$ being the common remainder. 

\begin{definition} Let $\mathfrak Z := (\mathbb Z,+,\cdot, \mathbb Z)$. Define the \textit{\textbf{$\mathfrak m$-fragment}}\footnote{See \prettyref{§:kl}.} of similarity by
\begin{align*} 
	\mathfrak m := \{xm+r \mid 0\leq r < m\},
\end{align*} and
\begin{align*} 
	\uparrow_ \mathfrak Z^ \mathfrak m a :=\ (\uparrow_ \mathfrak Z a)\cap \mathfrak m,
\end{align*} and $\approx_\mathfrak m$ is defined as $\approx$ in \prettyref{d:approx} with $\uparrow$ replaced by $\uparrow^ \mathfrak m$.
\end{definition}

The next result shows that modular arithmetic corresponds to a simple fragment of similarity interpreted in the arithmetic setting (we omit $\mathfrak Z$):

\begin{theorem} $a\equiv_m b$ iff $a\approx_ \mathfrak m b$.
\end{theorem}
\begin{proof} An immediate consequence of
\begin{align*} 
	xm+r\in a\uparrow b \quad\Leftrightarrow\quad a\equiv_m b.
\end{align*}
\end{proof}

\subsection{Semigroups}\label{§:Semigroups}

In any semigroup $\mathfrak S=(S,\cdot,S)$ consisting of a set $S$ together with an associative binary relation $\cdot$ on $S$ and constants for each element $S$, recall Green's well-known relations at the center of semigroup theory (see e.g. \cite{Howie03}) given by
\begin{align*} 
	a\leqq_ \mathcal L b & \quad:\Leftrightarrow\quad \text{$a=cb$, for some $c$},\\
	a\leqq_ \mathcal R b & \quad:\Leftrightarrow\quad \text{$a=bc$, for some $c$},\\
	a\leqq_ \mathcal J b & \quad:\Leftrightarrow\quad \text{$a=cbd$, for some $c,d$},
\end{align*} and
\begin{align*} 
	a\equiv_ \mathcal L b & \quad:\Leftrightarrow\quad a\leqq_ \mathcal L b \quad\text{and}\quad b\leqq_ \mathcal L a,\\
	a\equiv_ \mathcal R b & \quad:\Leftrightarrow\quad a\leqq_ \mathcal R b \quad\text{and}\quad b\leqq_ \mathcal R a,\\
	a\equiv_ \mathcal J b & \quad:\Leftrightarrow\quad a\leqq_ \mathcal J b \quad\text{and}\quad b\leqq_ \mathcal J a.
\end{align*} We can reformulate the relations in terms of generalizations as
\begin{align} 
	a\leqq_ \mathcal L b \quad&\Leftrightarrow\quad xb\in a\uparrow b,\\
	a\leqq_ \mathcal R b \quad&\Leftrightarrow\quad bx\in a\uparrow b,\\
	\label{eq:J} a\leqq_ \mathcal J b \quad&\Leftrightarrow\quad xby\in a\uparrow b.
\end{align} This indicates that Green's relations are related to (a fragment of) similarity (cf. \prettyref{t:J}):

\begin{definition} Define the \textit{\textbf{$J$-fragment}} of similarity by
\begin{align*} 
	J := \{xay \mid a\in S\}
\end{align*} and
\begin{align*} 
	\uparrow^J_ \mathfrak S a := (\uparrow_ \mathfrak S a)\cap J,
\end{align*} and the \textit{\textbf{$J$-similarity relation}} $\approx_J$ is defined as $\approx$ in \prettyref{d:approx} with $\uparrow$ replaced by $\uparrow^J$.
\end{definition}

The next result shows that we can indeed capture Green's $\mathcal J$-relation via the $J$-fragment of similarity (we omit $\mathfrak S$):

\begin{theorem}\label{t:J} $a\equiv_ \mathcal J b$ iff $a\approx_J b$.
\end{theorem}
\begin{proof} $(\Rightarrow)$ We first notice that the assumption $a\equiv_ \mathcal J b$ implies
\begin{align}\label{eq: 240304-xay_xby} 
	xay,xby\in a\uparrow^J b
\end{align} and thus
\begin{align*} 
	(\uparrow^J a)\cup (\uparrow^J b)\neq \{\;\}.
\end{align*} Hence, to prove $a\lesssim_J b$ we need to show that $a\uparrow^J b$ is $b$-maximal. Suppose there is some $c\in M$ such that
\begin{align}\label{eq: 240304-abc} 
	a\uparrow^J b\subsetneq a\uparrow^J c.
\end{align} Then, by \prettyref{eq: 240304-xay_xby} we have
\begin{align*} 
	xay,xby\in a\uparrow^J c,
\end{align*} which implies
\begin{align*} 
	c\leqq_ \mathcal J a,b.
\end{align*} By \prettyref{eq: 240304-abc}, there is some
\begin{align*} 
	xdy\in (a\uparrow^J c)\setminus (\uparrow^J b),
\end{align*} which implies
\begin{align*} 
	a,c\leqq_ \mathcal J d \quad\text{whereas}\quad b\not\leqq_ \mathcal J d,
\end{align*} a contradiction to $a\equiv_ \mathcal J b$. This shows $a\lesssim_J b$, and an analogous argument shows $b\lesssim_J a$, which in total yields $a\approx_J b$.

$(\Leftarrow)$ The assumed $a$-, and $b$-maximality of $a\uparrow^J b$ implies
\begin{align*} 
	a\uparrow^J b =\ \uparrow^J a \supseteq \{xay\} \quad\text{and}\quad a\uparrow^J b =\ \uparrow^J b \supseteq \{xby\},
\end{align*} which yields
\begin{align*} 
	xay,xby\in a\uparrow^J b.
\end{align*} Now apply \prettyref{eq:J}.
\end{proof}

\begin{fact} In analogy to \prettyref{t:J}, we can capture Green's $\mathcal L$-, and $\mathcal R$-relation via the $L$-, and $R$-fragment of similarity, where $L:=\{xa\mid a\in S\}$ and $R:=\{ax\mid a\in S\}$ and $\approx_L$ and $\approx_R$ are defined as $\approx$ with $\uparrow$ replaced by $\uparrow^L$ and $\uparrow^R$, respectively.
\end{fact}

\subsection{Groups}\label{§:Groups}

Let $\mathfrak G=(G,\cdot,1)$ be a group, that is, $\cdot$ is associative, 1 is the unit element satisfying $a1=1a=a$ for all $a\in G$, and every $a$ has an inverse $a^{-1}$ such that $aa^{-1}=a^{-1}a=1$. 

Recall that two elements $a$ and $b$ of a group are \textit{\textbf{conjugate}} --- in symbols, $a\leq_ \mathfrak c b$ --- iff there is an element $g$ in the group such that $a=gbg^{-1}$. We write $a\equiv_ \mathfrak c b$ in case $a\leq_ \mathfrak c b$ and $b\leq_ \mathfrak c a$. This induces an equivalence relation whose equivalence classes are called \textit{\textbf{conjugacy classes}} (see e.g. \cite[p.89]{Hungerford74}).

\begin{definition} Define the \textit{\textbf{$C$-fragment}} of similarity by
\begin{align*} 
	C := \left\{xax^{-1} \;\middle|\; a\in G\right\},
\end{align*} and
\begin{align*} 
	\uparrow_ \mathfrak G^C a :=\ (\uparrow_ \mathfrak G a)\cap C,
\end{align*} and the \textit{\textbf{$C$-similarity relation}} $\approx_C$ is defined as $\approx$ in \prettyref{d:approx} with $\uparrow$ replaced by $\uparrow^C$.
\end{definition}

The next result shows that we can capture the conjugacy relation via the $C$-fragment of similarity (we omit $\mathfrak G$):

\begin{theorem} $a\approx_C b$ iff $a\equiv_ \mathfrak c b$.
\end{theorem}
\begin{proof} $(\Rightarrow)$ We have $a\lesssim_C b$ iff either $(\uparrow^C a)\cup (\uparrow^C b)=\{\;\}$ or $a\uparrow^C b$ contains a non-trivial generalization and is $b$-maximal. Since $\uparrow^ C a$ always contains $xax^{-1}$, we certainly\todo{only if there is $b$ such that $xax^{-1}\not\in\ \uparrow^C b$} have $(\uparrow^C a)\cup (\uparrow^C b)\neq \{\;\}$, which means that $a\uparrow^ C b$ must contain a non-trivial generalization. This implies the existence of some $xcx^{-1}\in a\uparrow^ C b$, for some element $c$, which is equivalent to
\begin{align*} 
	a=gcg^{-1} \quad\text{and}\quad b=hch^{-1},\quad\text{for some elements $g,h$}.
\end{align*} The equivalence
\begin{align*} 
	b=hch^{-1} \quad\Leftrightarrow\quad c=h^{-1}bh
\end{align*} shows
\begin{align*} 
	a=gcg^{-1}=(gh^{-1})b(hg^{-1})=(gh^{-1})b(gh^{-1})^{-1}
\end{align*} and thus $a\leq_ \mathfrak c b$.

$(\Leftarrow)$ We assume $a\leq_ \mathfrak c b$, which is equivalent to $a=gbg^{-1}$ for some element $g$. Hence, $xbx^{-1}\in a\uparrow^C b$. It remains to show that $a\uparrow^C b$ is $b$-maximal. If $c$ is such that $a\uparrow^C b\subseteq a\uparrow^C c$, then in particular $xbx^{-1}\in a\uparrow^C c\subseteq\ \uparrow^C c$, and hence $c=gbg^{-1}$ for some $g$ or, equivalently, $b=g^{-1}cg$. For every $xdx^{-1}\in a\uparrow^C c$, we have $xdx^{-1}\in a\uparrow^C b$ since $c=hdh^{-1}$ and $b=g^{-1}cg$ imply $b=(g^{-1}h)d(h^{-1}g)$. We have thus shown $a\uparrow^C c\subseteq a\uparrow^C b$ which implies the $b$-maximality of $a\uparrow^C b$.
\end{proof}

We now turn our attention to normal subgroups and factor groups. Recall that a subgroup $\mathfrak N=(N,\cdot,1)$ of $\mathfrak G=(G,\cdot,1)$ is a \textit{\textbf{normal subgroup}} iff $aN=Na$ for all $a\in G$ (see e.g. \cite[Definition 5.2]{Hungerford74}). Let $\mathfrak N=(N,\cdot,1)$ be a normal subgroup of $\mathfrak G$, and let $\mathfrak G/ \mathfrak N=(G/N,\cdot,N)$ denote the \textit{\textbf{factor group}} of $\mathfrak G$ with respect to $\mathfrak N$ consisting of the \textit{\textbf{cosets}} $aN=\{ag\mid g\in N\}$, $a\in G$, in $G/N$ where multiplication is given by
\begin{align}\label{eq:aNbN} 
	(aN)(bN)=(ab)N.
\end{align}

Interestingly, elements in $G$ and cosets in $G/N$ have the same generalizations:

\begin{lemma}\label{l:a_aN} $\uparrow_ \mathfrak G a =\ \uparrow_{ \mathfrak G/ \mathfrak N}aN$.
\end{lemma}
\begin{proof} We first show
\begin{align} 
	\label{eq:s=s} s^ \mathfrak G( \textbf{o})N = s^{ \mathfrak G/ \mathfrak N}( \textbf{o}N),
\end{align} for every term $s$ and sequence of elements $\textbf{o}\in G^{r(s)}$, by structural induction on the shape of $s$. The bases cases $s=1$ and $s=x$ are trivial. If $s=rt$, for some terms $r$ and $t$, then
\begin{align*} 
		s^ \mathfrak G( \textbf{o})N = r^ \mathfrak G( \textbf{o})t^ \mathfrak G( \textbf{o})N \stackrel{\prettyref{eq:aNbN}}= (r^ \mathfrak G( \textbf{o})N)(t^ \mathfrak G( \textbf{o})N) \stackrel{IH}= r^{ \mathfrak G/ \mathfrak N}( \textbf{o}N) t^{ \mathfrak G/ \mathfrak N}( \textbf{o}N) = s^{ \mathfrak G/ \mathfrak N}( \textbf{o}N).
\end{align*} 

Now, a term $s$ is in $\uparrow_ \mathfrak G a$ iff $a=s^ \mathfrak G( \textbf{o})$, for some $\textbf{o}\in G^{r(s)}$, which by \prettyref{eq:s=s} means that $aN=s^{ \mathfrak G/ \mathfrak N}( \textbf{o}N)$, which is equivalent to $s$ being in $\uparrow_{ \mathfrak G/ \mathfrak N}aN$.
\end{proof}

\begin{theorem} $(\mathfrak G,\mathfrak G/ \mathfrak N) \models a\approx aN$.
\end{theorem}
\begin{proof} An immediate consequence of \prettyref{eq:=} and \prettyref{l:a_aN}.
\end{proof}

\section{Logical interpretation}\label{§:Logical_Interpretation}

In this section, we show how the purely algebraic notion of similarity can be naturally interpreted within the \textit{logical} setting of first-order logic via the well-known concept of a model-theoretic type. For this, we associate with every $L$-term $s(\textbf{y})$ a \textit{\textbf{g-formula}} of the form
\begin{align*} 
	\alpha_{s(\textbf{y})}(x) :\equiv (\exists \textbf{y})(x=s(\textbf{y})).
\end{align*} We denote the set of all g-formulas over $L$ by $g\text-Fm_L$.

Now that we have defined g-formulas, we continue by translating sets of generalizations into sets of g-formulas via a restricted notion of the well-known model-theoretic types (see e.g. \cite[§7.1]{Hinman05}). Define the \textit{\textbf{g-type}} of an element $a\in \mathbb A$ by
\begin{align*} 
	g\text-Type_ \mathfrak A(a) := \left\{\alpha\in g\text-Fm_L \;\middle|\; \mathfrak A\models \alpha(a)\right\}.
\end{align*}

We have the following correspondence between sets of generalizations and g-types:
\begin{align*} 
	s\in\ \uparrow_ \mathfrak A a \quad\Leftrightarrow\quad \alpha_s\in g\text-Type_ \mathfrak A(a).
\end{align*} This is interesting as it shows that algebraic generalizations, which are motivated by simple examples, have an intuitive logical meaning. The expression ``$a$ is similar to $b$'' can thus be reinterpreted from a logical point of view by defining the relation $\approx_T$ in the same way as $\approx$ with $\uparrow$ replaced by $g\text-Type$. We then have the logical characterization of similarity in terms of model-theoretic types via
\begin{align*} 
	\mathfrak{(A,B)} \models a\approx b \quad\Leftrightarrow\quad \mathfrak{(A,B)} \models a\approx_T b.
\end{align*}

Analogous procedures yield logical interpretations of the fragments in \prettyref{§:kl}.

\section{Towards applications to TCS and AI}\label{§:TCS_and_AI}

In this last semi-technical section, I want to sketch some potential applications of similarity to theoretical computer science and artificial intelligence.

Mathematically speaking, similarity is a structure-preserving binary relation between algebras, which can be used to identify similar objects across different domains. This can be exploited in any situation where it is desirable to \textit{generate} novel objects from existing ones in a systematic and approximative manner. This is similar in spirit to Generative AI,\footnote{\url{https://en.wikipedia.org/wiki/Generative_artificial_intelligence}} which has recently gained popularity from practically useful systems like OpenAI's ChatGPT\footnote{\url{https://de.wikipedia.org/wiki/ChatGPT}} and other so-called large language models (LLMs).\footnote{\url{https://en.wikipedia.org/wiki/Large_language_model}} 

\subsection{Program and theory synthesis}

Re-usability is key to software engineering and --- due to the deep connection between programs and proofs witnessed by the Curry-Howard correspondence --- to automated deduction and theorem proving. Since problem solving and theory (or program) generation are two sides of the same coin, it is of practical relevance to have mathematical tools available for the re-use and transfer of existing libraries of code in both engineering and math. The framework of similarity put forward in this paper may contribute in the future to that problem.

More specifically, a program $P$ written in a given programming language is a syntactic object with a semantics --- its behavior --- which we will denote by $P^\omega$. As soon as we have an \textit{algebra of programs} at hand, we can instantiate the framework of this paper to \textit{automatically} obtain a notion of program similarity! 

More formally, if $\mathfrak P$ and $\mathfrak R$ are two program algebras, and $P$ and $R$ are programs in $\mathfrak P$ and $\mathfrak R$, respectively, we can study the similarity relation between programs and their semantics
\begin{align*} 
	( \mathfrak{P,R}) \models P\approx R \quad\text{and}\quad ( \mathfrak{P,R}) \models P^\omega\approx R^\omega.
\end{align*} 

By definition, two syntactically and semantically similar programs yield similar results when executed. That is, if $P$ satisfies a specification $\psi$ and $P$ is similar to $R$, then $R$ satisfies a specification $\varphi$ similar to $\psi$.

\subsection{Neural-symbolic integration}

Today, ``good old fashioned'' symbolic AI,\footnote{\url{https://en.wikipedia.org/wiki/GOFAI}} based on logic (programming) \cite{Apt90,Lloyd87}, and ``modern'' connectionist subsymbolic AI, based on artificial neural networks \cite{ROJAS96} and deep learning \cite{LeCun15} (and see e.g. \cite{Goodfellow16}), are separated fields of AI-research. Both worlds have their strengths and weaknesses. Logical formalisms can be interpreted by humans and have a clear formal semantics which is missing for neural nets. Connectionist systems, on the other hand, have a remarkable noise-tolerance and learning capability which is missing for logical formalisms (a notable exception is inductive logic programming \cite{Muggleton91}). 

Neural-symbolic integration tries to unify both approaches (see e.g. \cite{Garcez15,Garcez02,Garcez09,DeRaedt20,Valiant08}). Compared to the field's short existence, its successes are remarkable and can be found in various fields such as bioinformatics, control engineering, software verification and adaptation, visual intelligence, ontology learning, and computer games \cite{Borges11,dePenning11,Hitzler05}.

An algebraic notion of similarity, as developed in this paper, can provide a mechanism for establishing correspondences between logic programs and neural nets in the spirit of neural-symbolic AI. 

More precisely, given an algebra of logic programs $\mathfrak P$ and an algebra of neural nets $\mathfrak N$, and a concrete program $P$ in $\mathfrak P$ and a net $N$ in $\mathfrak N$, the framework developed in this paper tells us when $P$ and $N$ are similar, in symbols
\begin{align*} 
	( \mathfrak{P,N}) \models P\approx N.
\end{align*} 

Now if we start with a net $N$, for which we have no symbolic explanation, we could try to compute one or more logic programs similar to $N$ serving as (approximate) explanations of the behavior of $N$. This could help with the difficult task of explaining the meaning of a neural net in symbolic terms as part of Explainable AI \cite{Doran18,Heder23}.

However, notice that for the framework to be applicable, algebras of logic programs and neural nets have to be introduced first. On the logic programming side, O'Keefe \cite{OKeefe85} was the first to propose an algebraic approach to logic programming, which has then been followed up by researchers interested in modular logic programming (see e.g. \cite{Brogi99,Bugliesi94}), and more recently by the author \cite{Antic14,Antic23-23,Antic21-2,Antic21-1}. A corresponding algebra of neural nets, on the other hand, appears to be missing and is a highly interesting line of future work. One approach is to reformulate neural nets as a form of neural-like logic programs \cite{Antic23-10} and to define an algebra of such programs similar to the one for ordinary programs.

\subsection{Analogical reasoning}

Lastly, I want to mention that similarity, as developed in this paper, can be used to define analogical proportions \cite{Antic22} of the form ``$a$ is to $b$ as $c$ is to $d$'' --- which are vital to analogical reasoning in AI (see e.g. \cite{Hofstadter95a,Prade21}) --- as demonstrated in \cite{Antic23-24}. This further emphasizes the fact that similarity is really at the core of analogy-making and thus at the core of cognition \cite{Gust08,Hofstadter01}.

\section{Conclusion and future work}\label{§:Conclusion}

This paper introduced from first principles an abstract algebraic and \textit{qualitative} notion of similarity within the general setting of universal algebra based on sets of generalizations motivated by the observation that these sets encode important properties of elements. It turned out that similarity defined in this way has appealing mathematical properties.

\subsection*{Outlook}

At the core of similarity is the set $\uparrow_ \mathfrak A a$ of all generalizations of an element $a$ in an algebra $\mathfrak A$. It appears challenging to compute $\uparrow_ \mathfrak A a$ even in simple concrete domains of infinite cardinality like the multiplicative algebra or the word domain. A reasonable starting point are unary algebras for which we have established, in the finite case, a close relationship to regular languages and finite automata in \prettyref{§:FUA}. 

We have seen in \prettyref{e:H_1} that the First \ref{t:FIT} and Second Isomorphism \prettyref{t:SIT} can in general not be generalized to homomorphisms. Are there algebras in which homomorphisms \textit{are} compatible with similarity and how can they be characterized? More generally speaking, what kind of mappings are compatible with similarity?


Another task is to study the $(k,\ell)$-fragments of \prettyref{§:kl} where generalizations are syntactically restricted. In \prettyref{§:kl}, we sketched the monolinear fragment where generalizations contain only a single occurrence of a single variable. Studying the linear fragment consisting of generalizations containing at most one occurrence of multiple variables in the abstract and in concrete algebras is one among many possible next steps.

We have seen in \prettyref{t:rst} that similarity is in general not transitive justified by a simple counterexample. In those algebras where similarity \textit{is} transitive, it forms an equivalence relation and enjoys all the nice properties that come with it. It is therefore desirable to have general criteria under which algebras induce a transitive similarity relation.


Another line of work is to study the connections between similarity and \textit{algebraic} anti-unification as recently introduced in \cite{Antic23-19}. A reasonable starting point is to study word similarity using Biere's \cite{Biere93} anti-unification algorithm for words.

Anti-unification \cite{Plotkin70,Reynolds70} is \textit{the} theory of generalization (cf. \cite{Cerna23}). The name is derived from the fact that computing the least general generalization of two terms is the dual of computing their \textit{unification}, an operation heavily studied in the literature with applications to automated theorem proving and logic programming (cf. \cite{Baader01}). The main focus of anti-unification is the computation of (complete sets of) least or minimally general generalizations. The problem of computing or representing the set of \textit{all} generalizations --- as used in this paper --- has, to the best of our knowledge, not been studied in anti-unification theory. In fact, Temur Kutsia --- an expert in the field of anti-unification --- thinks that existing tools cannot be adopted in a straightforward manner and that therefore novel tools will be needed for the computation of the set of all generalizations (recall that in \prettyref{§:FA} we have shown how to compute the set of $k$-generalizations in finite algebras).\footnote{Personal communication.}

A practically interesting domain of study relevant for applications in computational linguistics, biology, and AI in general, is the word domain. Again, computing the set of all generalizations of a given word appears challenging. Nonetheless, it is interesting to study similarity of words and how it is related to the numerous notions of word similarity in the literature (cf. \cite{Navarro01}).


Finally, it will be interesting to compare our notion of similarity to bisimilarity which is fundamental to theoretical computer science, logic, and mathematics (see e.g. \cite{Sangiorgi12}).

\section*{Acknowledgments}

I would like to thank the reviewers for their thoughtful and constructive comments, and for their helpful suggestions to improve the presentation of the article.

\section*{Compliance with Ethical Standards}

\subsection*{Ethical approval}

This article does not contain any studies with human participants performed by any of the authors.

\subsection*{Funding}

The authors declare that no funds, grants, or other support were received during the preparation of this manuscript.

\subsection*{Competing interests}

The authors declare that they have no conflict of interest.

\subsection*{Author contributions}

Single author.

\subsection*{Data availability statement}

The manuscript has no data associated.

\if\isdraft1\newpage\fi
\bibliographystyle{acm}
\bibliography{/Users/christianantic/Bibdesk/Bibliography,/Users/christianantic/Bibdesk/Publications_J,/Users/christianantic/Bibdesk/Publications_C,/Users/christianantic/Bibdesk/Preprints,/Users/christianantic/Bibdesk/Under_construction}
\if\isdraft1\newpage

\section{Equivalence and congruence relations}

\todo[inline]{Can every equivalence or congruence relation be modeled via a fragment of similarity?}



\section{Similarity-based homomorphisms and congruences}

\begin{definition} A \textit{\textbf{similarity-based homomorphism}} (or \textit{\textbf{s-homomorphism}}) is any mapping $\bullet: \mathfrak A\to \mathfrak B$ satisfying
\begin{align*} 
	\mathfrak A \models a\approx b \quad\Leftrightarrow\quad \mathfrak B \models a^\bullet\approx b^\bullet,\quad\text{for all $a,b\in \mathbb A$}.
\end{align*}
\end{definition}

\begin{definition} A \textit{\textbf{similarity-based congruence}} (or \textit{\textbf{s-congruence}}) on $\mathfrak A$ is an equivalence relation $\theta$ on $\mathbb A$ satisfying, for all $a,b,c,d\in \mathbb A$,
\begin{prooftree}
	\AxiomC{$a\approx b\quad a\theta c\quad b\theta d$}
	\RightLabel{.}
	\UnaryInfC{$c\approx d$}
\end{prooftree}
\end{definition}

\section{Analogies}

\begin{definition} An \textit{\textbf{analogy}} is any mapping $\bullet: \mathfrak A\to \mathfrak B$ satisfying
\begin{align*} 
	\mathfrak{(A,B)} \models a\approx a^\bullet,\quad\text{for all $a\in \mathbb A$}.
\end{align*} A \textit{\textbf{bianalogy}} is a bijective analogy, a \textit{\textbf{homoanalogy}} is an analogy which is a homomorphism, and an \textit{\textbf{isoanalogy}} is an analogy which is an isomorphism as well.
\end{definition}

\begin{example} 
\todo[inline]{}
\end{example}

\section{Linear multiplicative natural number similarity}

We work in $\mathfrak M=(\mathbb N_2,\cdot, \mathbb N_2)$.

\begin{fact}\label{f: 230903-a|b_uparrow^l} If $a\mid b$ then $\uparrow^l a\subseteq\ \uparrow^l b$.
\end{fact}
\begin{proof} For every linear generalization $a'z_1\ldots z_n$ of $a$, since $a\mid b$ and $a'\mid a$, we have $a'\mid b$; moreover, since $a\mid b$, we have $len(a)\leq len(b)$, which shows that $a'z_1\ldots z_n\in\ \uparrow^l b$. 
\end{proof}

We cannot expect the converse of \prettyref{f: 230903-a|b_uparrow^l} to hold since we can always choose a prime $p$ large enough so that
\begin{align*} 
	\uparrow^l p=\{x\}\subseteq\ \uparrow^l b \quad\text{and}\quad p>b.
\end{align*} Hence, we have shown:

\begin{fact} $\uparrow^l a\subseteq\ \uparrow^l b$ does \textit{not} imply $a\mid b$.
\end{fact}

Moreover, we cannot expect \prettyref{f: 230903-a|b_uparrow^l} to hold beyond the linear case since, for example,
\begin{align*} 
	4\mid 8 \quad\text{and}\quad x^2\in\ \uparrow 4 \quad\text{whereas}\quad x^2\not\in\ \uparrow 8.
\end{align*}

\begin{lemma}\label{l:uparrow^l_a|b} $a\uparrow^l b$ is $b$-maximal iff $a\mid b$.
\end{lemma}
\begin{proof} The direction from right to left is an immediate consequence of \prettyref{f: 230903-a|b_uparrow^l}. 

For the other direction, it suffices to show 
\begin{align*} 
	a\nmid b \quad\text{implies}\quad \uparrow^l a\not\subseteq\ \uparrow^l b.
\end{align*} In case $a\nmid b$, there is some prime power $p^k$ that divides $a$ but not $b$, which means that
\begin{align*} 
	p^kz\in\ \uparrow^l a \quad\text{whereas}\quad p^kz\not\in\ \uparrow^l b.
\end{align*}
\end{proof}

\begin{proposition}\label{p:a_lesssim_b_iff_a|b} $\mathfrak M\models_l a\lesssim b$ iff $a\mid b$. This implies: $\mathfrak M\models_l a\approx b$ iff $a=b$.
\end{proposition}
\begin{proof} An immediate consequence of \prettyref{l:uparrow^l_a|b}.
\end{proof}

\section{Motivation}

Gehe auf \cite{Yao00} ein --- vielleicht bekommst du dort eine plausible Motivation fuer das Paper!\todo{}

\section{Morphisms}

Given that $a\approx b$, does there exist a homomorphism $\bullet$ such that $a^\bullet=b$? One requirement is $$\uparrow^{at} a\subseteq\ \uparrow^{at} b.$$ Wenn das nicht gegeben ist, braucht es eine andere Bedingung, die weaker ist --- hier kommt Similarity ins Spiel. Motiviert ist das Ganze aus der Praxis im Programmieren, wo wir ein Modul $M$ durch ein Modul $N$ ersetzen wollen, obwohl $\uparrow^{at} M\not\subsetneq\ \uparrow^{at}N$ gilt.

\section{Related work}\label{§:RW}

\todo[inline]{Erwaehne Galois-Verbindung wie von M. Kompatscher beschrieben!}

The idea of this paper to use generalizations to define similarity in the algebraic setting of universal algebra appears to be original.

There is a vast literature on similarity, most of which deals with \textit{quantitative} approaches where similarity is numerically measured in one way or another. A notable exception is \cite{Yao00} where qualitative judgements of similarity are interpreted through statements of the form ``$a$ is more similar to $b$ than $c$ to $d$.'' In \cite{Badra18}, the authors analyze the role of qualitative similarity in analogical transfer.

Anti-unification \cite{Plotkin70,Reynolds70} is \textit{the} theory of generalization (cf. \cite{Cerna23}). The name is derived from the fact that computing the least general generalization of two terms is the dual of computing their \textit{unification}, an operation heavily studied in the literature with applications to automated theorem proving and logic programming (cf. \cite{Baader01}). The main focus of anti-unification is the computation of (complete sets of) least or minimally general generalizations. The problem of computing or representing the set of \textit{all} generalizations has, to the best of our knowledge, not been studied in anti-unification theory. In fact, Temur Kutsia (personal communication) --- an expert in the field of anti-unification --- thinks that existing tools cannot be adopted in a straightforward manner and that therefore novel tools will be needed.

In \prettyref{§:FUA} we have seen that in finite unary algebras, sets of generalizations form regular languages. Something similar in spirit has been done in \cite{Cerna20} where the solution set to an idempotent anti-unification problem is represented in the form of a regular tree language.


\section{Primitive positive formulas}

Der TOCL-Reviewer schreibt:
\begin{quote} 
	The notion of similarity being explored is the following. We associate with every element $a$ in an algebra $\mathfrak A$ the collection of terms which (with free variables suitably instantiated) take value $a$. In logical terms, this is simply the collection of primitive positive (PP) formulas that are true of $a$. This logical formulation is not explored in the present submission. Indeed, the main theorems --- Theorems X and Y --- are immediate consequences of the standard logical isomorphism theorems when formulated in these terms. The formulation in terms of PP formulas might unlock connections to the vast literature on PP formulas, including the work on CSP and homomorphisms. This is not pursued in the present paper.
\end{quote}

\section{Downarrow}

\begin{definition} For an $L$-term $s$, define
\begin{align*} 
	\downarrow_ \mathfrak A s := \left\{s^\mathfrak A( \textbf{o})\in \mathbb A \;\middle|\; \textbf{o}\in \mathbb A^{r(s)}\right\},
\end{align*} extended to a set of $L$-terms $S$ by
\begin{align*} 
	\downarrow_ \mathfrak A S := \bigcap_{s\in S}\downarrow_ \mathfrak A s,
\end{align*} and further extended to pairs of $L$-algebras by
\begin{align*} 
	\downarrow_ \mathfrak{(A,B)} s := (\downarrow_ \mathfrak A s)\cup (\downarrow_ \mathfrak B s)
\end{align*} and
\begin{align*} 
	\downarrow_ \mathfrak{(A,B)} S := (\downarrow_ \mathfrak A S)\cup (\downarrow_ \mathfrak B S).
\end{align*}
\end{definition}

\begin{fact} We have
\begin{align*} 
	S\subseteq\ \uparrow_ \mathfrak A a \quad\Leftrightarrow\quad a\in\ \downarrow_ \mathfrak A S,
\end{align*} and $S$ is a characteristic set of generalizations of $a\lesssim b$ in $\mathfrak{(A,B)}$ iff $a\in\ \downarrow_ \mathfrak A S$ and $\downarrow_\mathfrak B b=\{b\}$\todo{$\downarrow_\mathfrak B S$?}.
\end{fact}

\section{Similarity in finite unary algebras}

\begin{example} In the monounary algebra $\mathfrak A$ given by
\begin{center}
\begin{tikzpicture} 
    \node (1) {1};
    \node (2) [right=of 1,xshift=1.5cm] {2};
    \draw[->] (1) to [edge label'={$S$}][loop] (1);
    \draw[<->] (1) to [edge label={$T$}] (2);
    \draw[->] (2) to [edge label'={$S$}][loop] (2);
\end{tikzpicture}
\end{center} we have
\begin{align*} 
    1\approx 2.
\end{align*} Now consider the monounary algebra $\mathfrak B$ given by
\begin{center}
\begin{tikzpicture} 
    \node (3) {3};
    \draw[->] (3) to [edge label'={$S,T$}][loop] (3);
\end{tikzpicture}
\end{center} Since
\begin{align*} 
    \emptyset\neq 1\uparrow 3 = 2\uparrow 3,
\end{align*} we have
\begin{align*} 
    1,2\approx 3
\end{align*} and thus
\begin{align*} 
    \mathfrak A\approx \mathfrak B.
\end{align*}
\end{example}

\begin{example} Let $\mathfrak N=(\mathbb N,S)$ and $\mathfrak B=(\{0,1\},\neg)$. We have
\begin{align*} 
    a\lesssim_ \mathfrak{(N,B)} b,\quad\text{for all $b\in\{0,1\}$},
\end{align*} since
\begin{align*} 
    \uparrow_ \mathfrak B b= \mathbb N,
\end{align*} and
\begin{align*} 
    b\not\lesssim_ \mathfrak{(B,N)} a,\quad\text{for all $a\in \mathbb N$},
\end{align*} since
\begin{align*} 
    b\uparrow_ \mathfrak{(B,N)} a\subsetneq b\uparrow_ \mathfrak{(B,N)} a+c,\quad\text{for all $c\in \mathbb N$}.
\end{align*} Moreover, we have
\begin{align*} 
    S\lnsim \neg.
\end{align*} This shows
\begin{align*} 
    \mathfrak N\lnsim \mathfrak B.
\end{align*}
\end{example}

\begin{example} Consider the monounary algbras $\mathfrak A=([1,5],S)$ and $\mathfrak B=([6,9],T)$ given by
\begin{center}
\begin{tikzpicture} 
    \node (1) {1};
    \node (2) [right=of 1] {2};
    \node (3) [right=of 2] {3};
    \node (*) [right=of 3] {};
    \node (4) [above=of *] {4};
    \node (5) [below=of *] {5};
    \draw[->] (1) to [edge label={$S$}] (2);
    \draw[->] (2) to [edge label={$S$}] (3);
    \draw[->] (3) to [edge label={$S$}] (4);
    \draw[->] (4) to [edge label={$S$}] (5);
    \draw[->] (5) to [edge label={$S$}] (3);

    \node (6) [right=of *,xshift=1cm] {6};
    \node (7) [right=of 6] {7};
    \node (**) [right=of 7] {};
    \node (8) [above=of **] {8};
    \node (9) [below=of **] {9};
    \draw[->] (6) to [edge label={$T$}] (7);
    \draw[->] (7) to [edge label={$T$}] (8);
    \draw[->] (8) to [edge label={$T$}] (9);
    \draw[->] (9) to [edge label={$T$}] (7);
\end{tikzpicture}
\end{center} The element $2$ is unique as we have
\begin{align*} 
    2\not\approx b,\quad\text{for all $b\in [6,9]$}.
\end{align*} We only have
\begin{align*} 
    2\lnsim 7,8,9.
\end{align*} For every other element there is an element in the opposite algebra similar to it:
\begin{align*} 
    1\approx 6 \quad\text{and}\quad 3,4,5\approx 7,8,9.
\end{align*} We thus have
\begin{align*} 
    [6,9]\lnsim_ \mathfrak{(B,A)} [1,5].
\end{align*} What about the unary functions $S$ and $T$? For the element $1$ we have no element $b\in [6,9]$ satisfying $S1\approx_ \mathfrak{(A,B)} Sb$; on the other hand, for every $b\in [6,9]$ there is some $a\in [3,5]$ with $Sb\approx_ \mathfrak{(B,A)} a$ which in total yields
\begin{align*} 
    T\lnsim_ \mathfrak{(B,A)} S.
\end{align*} Hence,
\begin{align*} 
    \mathfrak B\lnsim \mathfrak A,
\end{align*} which coincides with our intuition that there is a part of $\mathfrak A$ similar to $\mathfrak B$ but not vice versa.
\end{example}

\section{Term similarity}

In this section, we collect some simple observations regarding term similarity. We work here in the free term algebra.

\begin{fact} We have $x\lesssim t$, for every variable $x$ and term $t$. Consequently, $x\approx y$ holds for all variables $x$ and $y$, that is, all variables are similar.
\end{fact}
\begin{proof} A direct consequence of the fact that $x\uparrow t=\{x\}$, for every term $t$.
\end{proof}

\begin{fact} Two ground terms are similar iff they are identical.
\end{fact}
\begin{proof} For a ground term $s$, we have $s\in\ \uparrow s$, which means that for every ground term $t$, $s$ has to be in $s\uparrow t$ which can only hold iff $s=t$.
\end{proof}

\begin{fact} Two abstract\todo{def} terms are similar iff they are identical modulo the renaming of variables.
\end{fact}

\begin{example} $f(x,x)\approx f(y,y)$ whereas $f(x,y)\lnsim f(x,x)$, that is, $f(x,x)\not\approx f(x,y)$.
\end{example}

\begin{fact} If $s$ is an instance of $t$, then $t\lesssim s$.
\end{fact}

\begin{example} $f(x)\lnsim f(a)$.
\end{example}



\section{Set similarity}

\begin{fact} The empty set is always similar to itself for arbitrary pairs of set algebras.
\end{fact}
\begin{proof} Characteristically justified by $X\cap X^c$.
\end{proof}

\begin{fact} Two universes are always similar.
\end{fact}
\begin{proof} Characteristically justified by $X\cup X^c$.
\end{proof}

\section{}

\begin{fact} Falls es ein $s\in a\uparrow b$\footnote{Beachte, dass dann $s\in a\Uparrow b$.} gibt, sodass $\downarrow s = \{a,b\}$, dann $a\approx b$.
\end{fact}

\section{Maximal similarity and interfield theories}

Maximal similarity wird in \cite{Greiner88a} erwaehnt. Darin werden auch \textit{Interfield theories} \cite{Darden77} referenziert, die dazu dienen, zwei Theorien zu verknuepfen (so, wie ich das vor Augen hatte mit einer ``Bridge-Theorie'' $B$, die zwei Theorie $S$ und $T$ verknuepft).

\section{Similarity and bisimilarity}

The following counterexample shows that similarity and bisimilarity are different concepts:

\begin{example}\label{e: 240305-similarity_neq_bisimilarity} Consider the two labelled transitions systems $\mathfrak S$ and $\mathfrak T$ given respectively by
\begin{center}
\begin{tikzpicture} 
	\node (1) {1};
	\node (2) [right=of 1] {2};
	\node (3) [right=of 2] {3};
	\node (4) [right=of 3] {4};
	\node (5) [right=of 4] {5.};
	\draw[->] (1) to [edge label={$S$}, bend left] (2);
	\draw[->] (2) to [edge label={$T$}, bend left] (1);
	\draw[->] (3) to [edge label={$S$}] (4);
	\draw[->] (4) to [edge label={$T$}, bend left] (5);
	\draw[->] (5) to [edge label={$S$}, bend left] (4);
\end{tikzpicture}
\end{center} We assume here that we have already generalized the framework of analogical proportions to partial algebras. We have
\begin{align*} 
	\mathfrak S\lnsim \mathfrak T
\end{align*} since there is no element in $\mathfrak S$ similar to $3$. Bisimilarity, on the other hand, regards the states $1$ and $3$ as bisimilar.
\end{example}

\section{}

\begin{fact} If there is some $s\in a\uparrow b$ with $\downarrow s = \{a,b\}$,\footnote{Notice that we then have $s\in a\Uparrow b$.} then $a\approx b$.
\end{fact}

\fi
\end{document}